%% file: main.tex
\documentclass{article}

\usepackage{arxiv}

\usepackage[utf8]{inputenc} %
\usepackage[T1]{fontenc}    %
\usepackage{hyperref}
\usepackage{url}            %
\usepackage{booktabs}       %
\usepackage{amsfonts}       %
\usepackage{nicefrac}       %
\usepackage{microtype}      %
\usepackage{graphicx}
\usepackage[giveninits=true, backend=biber, style=ieee, sorting=none, minbibnames=5, maxbibnames=7, isbn=false,url=true,eprint=false]{biblatex}
\bibliography{references.bib}

\usepackage{caption}
\usepackage{subcaption}
\usepackage{multicol}

\usepackage{doi}
\usepackage{amsmath}
\usepackage{amstext}
\usepackage{amssymb}
\usepackage{amsbsy}
\usepackage{cleveref}
\usepackage{mathtools}
\usepackage{siunitx}

\usepackage{bm}
\usepackage{bbm}
\usepackage{enumitem}
\usepackage{float}
\usepackage{comment}
\usepackage{graphicx}
\usepackage[export]{adjustbox}
\usepackage[section]{placeins}

\usepackage[font=small,labelfont=bf]{caption}

\usepackage[acronym]{glossaries}

\newacronym{pls}{PLS}{Partial Least Squares}
\newacronym{cv}{CV}{Cross-Validation}
\newacronym{pca}{PCA}{Principal Component Analysis}
\newacronym{pcr}{PCR}{Principal Component Regression}
\newacronym{ols}{OLS}{Ordinary Least Squares}
\newacronym{cl}{CL}{Cycle Life}
\newacronym{rr}{RR}{Ridge Regression}
\newacronym{en}{EN}{Elastic Net}
\newacronym{snr}{SNR}{Signal-to-Noise Ratio}
\newacronym{svd}{SVD}{Singular Value Decomposition}
\newacronym{lasso}{lasso}{Least Absolute Shrinkage and Selection Operator}
\newacronym{lfp}{LFP}{Lithium Iron Phosphate}
\newacronym{rss}{RSS}{Residual Sum of Squares}
\newacronym{rmse}{RMSE}{Root-Mean-Square Error}
\newacronym{nrmse}{NRMSE}{Normalized-Root-Mean-Square Error}
\newacronym{mse}{MSE}{Mean Squared Error}
\newacronym{mae}{MAE}{Mean Absolute Error}
\newacronym{mape}{MAPE}{Mean-Absolute-Percentage Error}
\newacronym{si}{SI}{Supplementary Information}
\newacronym{ocv}{OCV}{Open-Circuit Voltage}

\newcommand{\normx}[2][2]{\lVert#2\rVert_#1}

\DeclarePairedDelimiter\abs{\lvert}{\rvert}
\makeatletter
\let\oldabs\abs
\def\abs{\@ifstar{\oldabs}{\oldabs*}}
\let\oldnorm\norm
\def\norm{\@ifstar{\oldnorm}{\oldnorm*}}

\newcommand\footnoteref[1]{\protected@xdef\@thefnmark{\ref{#1}}\@footnotemark}
\makeatother
\title{Interpretation of High-Dimensional Linear Regression: Effects of Nullspace and Regularization Demonstrated on Battery Data}

\date{}

\makeatletter
\let\titlecopy\@title
\makeatother

\author{\href{https://orcid.org/0000-0001-8767-4101}{\includegraphics[scale=0.06]{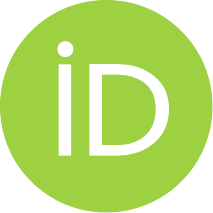}\hspace{1mm} Joachim Schaeffer} \\
    Technical University of Darmstadt \\ 
    Control and Cyber-Physical Systems Laboratory \\
    Karolinenpl. 5 \\
    Darmstadt, 64289, Germany
\And
\href{https://orcid.org/
0000-0003-1742-4750}{\includegraphics[scale=0.06]{orcid.pdf}\hspace{1mm} Eric Lenz} \\ 
Technical University of Darmstadt \\ 
Control and Cyber-Physical Systems Laboratory \\
Karolinenpl. 5 \\
Darmstadt, 64289, Germany
\And
\href{https://orcid.org/0000-0002-7066-3470}{\includegraphics[scale=0.06]{orcid.pdf}\hspace{1mm} William C. Chueh}\\
Stanford University \\
450 Jane Stanford Way \\ 
Stanford, 94305, CA, USA
\And
\href{https://orcid.org/0000-0002-8200-4501}{\includegraphics[scale=0.06]{orcid.pdf}\hspace{1mm}  Martin Z. Bazant} \\ 
Massachusetts Institute of Technology \\ 
77 Massachusetts Avenue \\ 
Cambridge, 02139, MA, USA
\And
\href{https://orcid.org/
0000-0002-9112-5946}{\includegraphics[scale=0.06]{orcid.pdf}\hspace{1mm}  Rolf Findeisen} \\ 
Technical University of Darmstadt \\ 
Control and Cyber-Physical Systems Laboratory \\
Karolinenpl. 5 \\
Darmstadt, 64289, Germany
\And
\href{https://orcid.org/
0000-0003-4304-3484}{\includegraphics[scale=0.06]{orcid.pdf}\hspace{1mm}  Richard D. Braatz} \\ 
Massachusetts Institute of Technology \\ 
77 Massachusetts Avenue \\ 
Cambridge, 02139, MA, USA \\
\texttt{braatz@mit.edu}
}
\renewcommand{\headeright}{}
\renewcommand{\undertitle}{}
\renewcommand{\shorttitle}{Interpretation of High-Dimensional Linear Regression: Effects of Nullspace and Regularization}

\hypersetup{
pdftitle={Interpretation of High-Dimensional Linear Regression: Effects of Nullspace and Regularization Demonstrated on Battery Data},
pdfsubject={stat.ML},
pdfauthor={Joachim Schaeffer, Eric Lenz, Martin Z. Bazant, Rolf Findeisen, Richard D. Braatz},
pdfkeywords={Data Matrix Nullspace, Regression Coefficients, Linear Regression},
}
\begin{document}
\maketitle

\input{content/00Abstract}

\keywords{Interpretable Machine Learning \and Linear Regression \and High Dimensions \and Nullspace \and Functional Data \and Regression Coefficients \and Lithium-Ion Batteries}

\input{content/01Introduction}

\input{content/02LinearRegression}

\input{content/03Methodology}

\input{content/04Examples}

\input{content/05Conclusion}

\section*{Code and Data Availability}
The code for this article is available in the corresponding GitHub repository, \textit{HDRegAnalytics}, \url{https://github.com/JoachimSchaeffer/HDRegAnalytics}. The repository contains the source code and notebooks to visualize the results. The repository contains a small subset of the \acrshort{lfp} dataset that was published with \cite{severson2019data} and is available at \url{https://data.matr.io/1/}.

\section*{Author Contributions}
Joachim Schaeffer: Conceptualization, Methodology, Software, Validation, Formal Analysis, Investigation, Data Curation, Writing -- original draft, Writing -- review \& editing, Visualization, Funding Acquisition; Eric Lenz: Formal Analysis, Writing -- review \& editing; William C. Chueh: Writing -- review \& editing; Martin Z. Bazant: Writing -- review \& editing; Rolf Findeisen: Resources, Writing -- review \& editing, Funding Acquisition;  Richard D. Braatz: Conceptualization, Methodology, Resources, Writing - original draft, Writing -- review \& editing, Supervision, Project Administration, Funding Acquisition. 

\section*{Acknowledgements and Funding}
Initial ideas for this work were conceptualized during Joachim Schaeffer's time at ETH Z\"{u}rich, for which we acknowledge financial support from the German Academic Exchange Service (DAAD) within the scholarship program for Master studies abroad.
The main work was carried out by Joachim Schaeffer at the Technical University of Darmstadt. The work was refined and extended during Joachim Schaeffer's time at the Massachusetts Institute of Technology, for which we acknowledge financial support by a fellowship within the IFI program of the German Academic Exchange Service (DAAD), funded by the Federal Ministry of Education and Research (BMBF). Furthermore, financial support is acknowledged from the Toyota Research Institute through the D3BATT Center on Data-Driven-Design of Rechargeable Batteries.

\newpage
\printbibliography

\clearpage
\appendix
\input{content/SupplementaryInformation}

\end{document}

%% file: content/00Abstract.tex
\begin{abstract}
    High-dimensional linear regression is important in many scientific fields. This article considers discrete measured data of underlying smooth latent processes, as is often obtained from chemical or biological systems.
   Interpretation in high dimensions is challenging because the nullspace and its interplay with regularization shapes regression coefficients. The data's nullspace contains all coefficients that satisfy $\mathbf{Xw}=\mathbf{0}$, thus allowing very different coefficients to yield identical predictions. 
    We developed an optimization formulation to compare regression coefficients and coefficients obtained by physical engineering knowledge to understand which part of the coefficient differences are close to the nullspace. 
    This nullspace method is tested on a synthetic example and lithium-ion battery data. The case studies show that regularization and z-scoring are design choices that, if chosen corresponding to prior physical knowledge, lead to interpretable regression results. Otherwise, the combination of the nullspace and regularization hinders interpretability and can make it impossible to obtain regression coefficients close to the true coefficients when there is a true underlying linear model.  
    Furthermore, we demonstrate that regression methods that do not produce coefficients orthogonal to the nullspace, such as fused lasso, can improve interpretability. In conclusion, the insights gained from the nullspace perspective help to make informed design choices for building regression models on high-dimensional data and reasoning about potential underlying linear models, which are important for system optimization and improving scientific understanding.
\end{abstract}

%% file: content/01Introduction.tex
\section{Introduction}
Many important regression problems have the dimensionality of the data $p$ much larger than the sample size $n$ \cite{buhlmann2011statistics, johnstone_stat_challenges_high_dimensional, hastie2009elements, kobak2020optimal, }. Consequently, $p \gg n$ and the matrix $\mathbf{X} \in \mathbb{R}^{n \times p}$  of predictors is ``wide''. This case arises, for example, in most spectroscopies, lithium-ion batteries \cite{ralbovsky2020towards, lavadeschafferbraatz2022}, brain imaging, and computational biology \cite{boulesteix2007partial, hastie2009elements, buhlmann2011statistics}. Classical literature on linear regression \cite{gross2003linear, montgomery2012linearregressionintroduction,seber2003linear} focuses mainly on the case where $p < n$ and mostly assumes full column rank; however, many linear regression methods work well with wide predictor matrices. While \gls{ols} is not defined for wide data matrices because $\mathbf{X}^\top\mathbf{X}$ is singular, the related minimum norm solution \cite{monticelli1999least} can be used instead. \gls{rr} and other shrinkage-based regression methods (e.g., \gls{lasso}, \gls{en}) do not suffer from this problem due to the penalty term that is added to the main diagonal of $\mathbf{X}^\top\mathbf{X}$. The fused \gls{lasso}, a generalization of the lasso, adds an L$_1$-norm penalty of adjacent regression coefficient differences to the objective function \cite{fused_lasso_tibshirani2007}. This additional penalty encourages piecewise constant regression coefficients, i.e., sparsity in regression coefficient differences. Thus it is required that the predictors can be ordered in some meaningful way. %
Latent variable methods such as \gls{pls} and \gls{pcr} are popular choices for high-dimensional regression in the chemometrics community. 

A key question is how to interpret high-dimensional linear regression results and the corresponding regression coefficients. In particular, how to reason about an underlying (linear) model for scientific insights and system optimization? Technically, regression coefficients for a linear model can be analyzed and compared to engineering or scientific expectations in terms of shape (e.g., peaks, plateaus, slopes), which is often done implicitly by engineers when looking at regression coefficients. However, as shown in this article, such an interpretation can lead to misleading conclusions.

This article develops a method, based on the nullspace of the predictor matrix $\mathcal{N}(\mathbf{X})$, for comparing coefficients obtained by different methods with each other for the case of high-dimensional data that was generated by a smooth latent process, also called {\it functional data} \cite{FunctionalDataAnalysis}.\footnote{Measured data from chemical and other systems often exhibit a certain degree of smoothness and can be considered to originate from discretized functions (similar to the assumptions made in \cite{functiononfunction}). The term {\it smoothness}, as used in this article, refers to data in which neighboring values are linked to each other to some extent, are not too different from one another, and there exists an underlying function that is differentiable once or multiple times.} %
We use the fact that $\mathcal{N}(\mathbf{X})$ consists of all solutions to $\mathbf{X}\mathbf{w} = 0$ and thus the predictions do not change when adding a vector of the nullspace to the regression coefficients $\mathbf{X}(\boldsymbol{\beta} + \mathbf {w}) = \mathbf{X}\boldsymbol{\beta}$. The nullspace and its interplay with regularization significantly influence the shape of the regression coefficients. Therefore, an understanding of the effect of the nullspace is needed for interpretation and scientific understanding. Our objective is to support such an understanding with this article.

The next section briefly introduces the key linear regression methods used in this article. Then the nullspace approach is derived. Subsequently, case studies are presented on fully synthetic data, lithium-ion battery data with two different synthetic linear responses, and the measured nonlinear cycle life response \cite{severson2019data}. The conclusion section summarizes the key learnings. All code and data used in this article are open-source and open-access, allowing the reproduction of results.

%% file: content/02LinearRegression.tex
\section{Motivation and Linear Regression}
\label{sec:linreg}
Linear, static models, assuming mean-centered data, have the general form
\begin{equation}
\mathbf{y} =
    \mathbf{X}\boldsymbol{\beta^*} + \boldsymbol{\epsilon}
    \label{eq:lin_model}
\end{equation}
where the input data matrix $\mathbf{X} \in \mathbb{R}^{n \times p}$, $n$ is the number of observations, $p$ is the number of predictors, and we assume that $p \gg n$. Our work is motivated by measurements of chemical or biochemical systems, i.e., discrete, noisy measurements of an assumed smooth underlying process.  Consequently, we assume a latent model structure, i.e., $\mathbf{X}$ (independently of $\mathbf{y}$) can be approximated in a lower dimensional space, and $\mathbf{X}$ is not sparse. Most of the analysis in this article is technically not limited to this assumption. However, the nullspace perspective is motivated by a latent model structure and the high multicollinearity of columns that arises from functional data. The coefficients $\boldsymbol{\beta} \in \mathbb{R}^{p}$ contain the relation between $\mathbf{X}$ and $\mathbf{y}$. 
The errors $\boldsymbol{\epsilon} \in \mathbb{R}^{p}$ are assumed to be homoscedastic, to have zero means, and to be uncorrelated. 

Linear regression denotes statistical methods to determine $\boldsymbol{\hat\beta}$ from data $\mathbf{X}$ and $\mathbf{y}$ minimizing the error $\boldsymbol{\hat\epsilon}$ concerning a defined measure of the error,
\begin{equation}
\mathbf{y} = \mathbf{\hat{y}} + \boldsymbol{\hat\epsilon} = \mathbf{X}\boldsymbol{\hat\beta} + \boldsymbol{\hat\epsilon}.
\label{eq:regression_model}
\end{equation}
The objective of regression methods is to find a $\boldsymbol{\hat\beta}$ that yields predictions that are reasonably close to the predictions of $\boldsymbol{\beta}^*$ when applied to independent data, i.e., were not available during training.  
When a true underlying linear model exists, interpretation and scientific insights would be supported by achieving a different goal, which is to reconstruct the true coefficients, i.e., $\boldsymbol{\hat\beta} = \boldsymbol{\beta}^*$, where $\boldsymbol{\beta}^*$ are the true coefficients of the model. 

As shown in \cite{lavadeschafferbraatz2022}, often columns in high-dimensional functional data are correlated, and regularized regression will find a solution that is optimal for its objective function; however, the resulting regression coefficients can be visually very different from $\boldsymbol{\beta}^*$ due to the interplay of the regularization and the nullspace, $\mathcal{N}(\mathbf{X})$. 
Furthermore, in practice, $\boldsymbol{\beta}^*$ is not known and the true underlying system might be nonlinear, requiring a thorough understanding of the interplay of regularization and the nullspace to draw reasonable conclusions about the underlying model.

Generally, the regression coefficients associated with $\boldsymbol{\hat\beta}^{\text{Model}}$ are random variables because $\mathbf{X}$ and $\mathbf{y}$ are realizations from a system that contains randomness (e.g., measurement errors, random system processes, etc.). One approach to model the regression coefficients probabilistically is Bayesian linear regression  which places a prior on the regression coefficients and yields their posterior distribution, conditioned on data, which can then be analyzed (e.g., see \cite{makalic2016high} for more information on Bayesian linear regression for high-dimensional data). 
While probabilistic modeling of regression coefficients is important, we focus on analyzing linear regression methods that do not model regression coefficients probabilistically because chemical engineers commonly use non-probabilistic models.
We use $\boldsymbol{\beta}^{\text{Model}}$ to denote that it is a realization of the random variable by the ``Model'' and specific training data. From here, we drop the ``hat'' notation because it is clear from the model name in the superscript that the coefficients were obtained by regression from data.

\Glsfirst{ols} regression estimates with the closed-form solution $\boldsymbol{\beta}^{\text{OLS}} = (\mathbf{X^{\top}X})^{-1}\mathbf{X^{\top}y}$ for the case $p<n$ have low bias and are optimal under the assumption of the Gauss-Markov theorem. However, the regression coefficients $\boldsymbol{\beta}$  have a very large variance if the condition number of $\mathbf{X}^\top\mathbf{X}$ is large, as is the case for many real-world data analytics problems, resulting in low prediction accuracy on unseen data.
\Glsfirst{rr} addresses this problem by adding the squared L$_2$-norm of the weights as a penalty to the least-squares objective \cite{hoerl1970ridge}:
\begin{equation}
    \min_{\boldsymbol{\beta}}  \|\mathbf{y}-\mathbf{X}\boldsymbol{\beta}\|_2^2 + \lambda\|\boldsymbol{\beta}\|_2^2 ,
    \label{eq:obj_rr}
\end{equation}
yielding the closed-form solution $\boldsymbol{\beta}^{\text{RR}} = (\mathbf{X^{\top}X + \lambda \mathbf{I}})^{-1}\mathbf{X^{\top}y}$. The regularization penalty adds to the main diagonal of $\mathbf{X^{\top}X}$ and ensures that the resulting matrix is also invertible in the case $p > n$. 
\gls{rr} improves the model's generalization by introducing a bias that reduces variance in the estimated parameters \cite{Zou2005}. 
For $p < n$ and $\lambda \to 0$, \gls{rr} converges to \gls{ols}. In the more general case, without making assumptions about the dimensionality and rank of the real matrix $\mathbf{X}$, \gls{svd}, $\mathbf{X} = \mathbf{U}\boldsymbol{\Sigma}\mathbf{V}^{\top}$, can be used to show that
\begin{equation}
    \boldsymbol{\beta}_0 = \lim_{\lambda \to 0} \boldsymbol{\beta}_{\lambda} = \mathbf{X}^{\dagger}y.
\end{equation}
The full derivation and further information can be found in the \gls{si}, Sec.\,\ref{SI:derivation} and \cite{kobak2020optimal}.
For the case $p > n$, the Moore-Penrose-Inverse $\mathbf{X}^{\dagger}$ can be written as
\begin{equation}
    \boldsymbol{\beta}_0 = \mathbf{X^{\top}}(\mathbf{XX^{\top})}^{-1}\mathbf{y}.
    \label{eq:MinNormSolution}
\end{equation}
This expression is known as the minimum norm solution (e.g., \cite{monticelli1999least}),
\begin{equation}
    \boldsymbol{\beta}_0 = \textrm{arg}\min_{\boldsymbol{\beta}} \left\{ \|\boldsymbol{\beta}\|_2^2 \ \middle\vert \ \|\mathbf{y} - \mathbf{X} \boldsymbol{\beta}\|_2^2 = \mathbf{0} \right\}.
\end{equation}

For any $\boldsymbol{\tilde\beta}$ that fulfills $\mathbf{X}\boldsymbol{\tilde\beta} = \mathbf{y}$ (i.e., regression coefficients that fit the data $\mathbf{X}$ and $\mathbf{y}$ perfectly including the noise), \eqref{eq:MinNormSolution} can be used to show that $\mathbf{X}(\boldsymbol{\tilde\beta}-\boldsymbol{\beta}_0) = \mathbf{0}$, and that
\begin{align}
    (\boldsymbol{\tilde\beta}-\boldsymbol{\beta}_0)^{\top}\boldsymbol{\beta}_0
    &= (\boldsymbol{\tilde\beta}-\boldsymbol{\beta}_0)^{\top} \mathbf{X}^{\top} (\mathbf{XX}^{\top})^{-1}\mathbf{y}\nonumber\\
    &= (\mathbf{X}(\boldsymbol{\tilde\beta}-\boldsymbol{\beta}_0))^{\top} (\mathbf{XX}^{\top})^{-1}\mathbf{y}\nonumber\\
    &= \mathbf{0} 
\end{align}
and consequently $(\boldsymbol{\tilde\beta}-\boldsymbol{\beta}_0) \perp \boldsymbol{\beta}_0$ which is equivalent to $\mathcal{N}(\mathbf{X}) \perp \boldsymbol{\beta}_0$ \cite{lns_stanford_lectureslides}. Thus there exists a set of regression coefficients $\mathbf{\mathcal{S}}$ that all fulfill $\mathbf{X}\boldsymbol{\tilde\beta}=\mathbf{y}$ with $\boldsymbol{\tilde\beta} \in \mathbf{\mathcal{S}}$. 

Most regularized regression methods solve an optimization of the form 
\begin{equation}
    \min_{\boldsymbol{\beta}}  \|\mathbf{y}-\mathbf{X}\boldsymbol{\beta}\|_2^2 + F(\boldsymbol{\beta}).
    \label{eq:obj_reg}
\end{equation}
Regularized methods trade the perfect fit to the training data against the objective of keeping regression coefficients small. This trade-off is seen in the objective used to define the regularization methods. The orthogonality of the regression coefficients to $\mathcal{N}(\mathbf{X})$ does not hold for arbitrary regularization terms $F(\boldsymbol{\beta})$. Orthogonality holds for \gls{rr}, because the regularization term in \eqref{eq:obj_rr} is always smaller for coefficients orthogonal to $\mathcal{N}(\mathbf{X})$. The \gls{pcr} coefficients are orthogonal to the $\mathcal{N}(\mathbf{X})$ because all eigenvectors of $\mathbf{X}^\top\mathbf{X}$ that correspond to nonzero eigenvalues are orthogonal to each other and to the nullspace. Similarly, the \gls{pls} coefficients are also orthogonal to the nullspace by construction. The pathological case of $\mathbf{y}$ being the nullvector must be excluded and is not relevant. Proofs for orthogonality between regression coefficients and nullspace for \gls{rr}, \gls{pcr}, and \gls{pls} are included in the \gls{si}, Sec.\,\ref{SI:orthogonality}.  %
However, regression coefficients obtained by the lasso and \gls{en} are not orthogonal to $\mathcal{N}(\mathbf{X})$ because of the L$_1$-norm. For more information on regularized high-dimensional regression, see \cite{lavadeschafferbraatz2022,Tibshirani1996}. 
Depending on the function $F\colon \mathbb{R}^p \to \mathbb{R}$, regression coefficients obtain different shapes. Usually, methods such as \gls{rr}, \gls{pcr}, and \gls{pls} yield solutions that are not sparse, which can make interpretation difficult. An alternative method is the \gls{lasso}, however, sparsity is often not a reasonable assumption for functional data.
A generalization of the lasso is
\begin{equation}
\min_{\boldsymbol{\beta}}  \frac{1}{2}\|\mathbf{y}-\mathbf{X}\boldsymbol{\beta}\|_2^2 + \lambda  \|\mathbf{D} \boldsymbol{\beta}\|_1 
\label{eq:gen_lasso}
\end{equation}
where choosing $\mathbf{D}$ as the identity matrix recovers the lasso. For 
\begin{equation}
    \mathbf{D_1} = 
    \begin{bmatrix}
        1 & -1 & 0 & \cdots  & 0\\
        0 & 1 & -1 & \ddots & \vdots\\
        \vdots & \ddots & \ddots & \ddots& 0 \\
        0 & \cdots & 0 & 1 & -1\\
    \end{bmatrix},
    \label{eq:d1}
\end{equation}
the resulting model is called the one-dimensional fused lasso which penalizes the L$_1$-norm of the regression coefficients as well as their differences, but thus requires predictors that can be ordered \cite{fused_lasso_tibshirani2007}, a characteristic of functional data \cite{FunctionalDataAnalysis}. The choice of $\mathbf{D}$ can incorporate expectations about the underlying model structure \cite{solution_path_gen_lasso}, and can thus yield models that should be interesting for many chemical engineering problems for its flexibility to incorporate assumptions, and its potential to yield easier-to-interpret regression coefficients. However, the authors are not aware that the fused lasso is currently being applied in the chemical engineering community, despite the popularity of the fused lasso in the statistics community. 
In the next section, we derive the nullspace method to compare the regression coefficients of different regularized models thoroughly.

%% file: content/03Methodology.tex
\section{Nullspace Method}
\label{sec:method}
The addition of a vector $\mathbf{v} \in \mathcal{N}(\mathbf{X})$, i.e., a vector in the nullspace, to any $\boldsymbol{\beta}$ yields coefficients with unchanged predictions. The vectors in the nullspace affect only the regularization term in the objective function.
We are interested in a method for understanding the effects of the nullspace when comparing different coefficients and how such a comparison can be used to reason about underlying relationships.
Consider a regularized regression model called A,
\begin{equation}
    \mathbf{\hat{y}}^\text{A} = \mathbf{X} \boldsymbol{\beta^\text{A}}, 
 \label{eq:model1}
 \end{equation}
where $\boldsymbol\beta^\text{A}$ is associated with method A. For any vector $\mathbf{v}$ in the nullspace, this equation implies that
\begin{equation}
    \mathbf{\hat{y}}^\text{A}
 = \mathbf{X} (\boldsymbol{\beta}^\text{A} + \mathbf{v}).
\end{equation}
We want to compare the regression coefficients $\boldsymbol{\beta}^\text{A}$ with other coefficients $\boldsymbol{\beta}^\text{B}$. The coefficients $\boldsymbol{\beta}^\text{B}$ can either be another estimator obtained by another regression method or instead be chosen for engineering or scientific reasons (e.g., constant regression coefficients).
Thus we propose finding coefficients $\mathbf{v}^* \in \mathcal{N}(\mathbf{X})$ that are closest to the difference between the coefficients under comparison  $\boldsymbol\beta_\Delta = \boldsymbol{\beta^\text{A}} - \boldsymbol{\beta^\text{B}}$. This approach can be formalized by
\begin{subequations}
  \label{eq:opt_simple_orig}
    \begin{alignat}{2}
        &\!\min_{\mathbf{v}}  \ \normx[2]{
\boldsymbol\beta_\Delta +\mathbf{v}}^2  \\
        \label{eq:opt_simple2}
        &\text{subject to } \mathbf{Xv} =\mathbf{0},
    \end{alignat}
\end{subequations}
This optimization is a convex quadratic program with linear constraints. The solution is the projection of $\boldsymbol\beta_\Delta$ onto the nullspace,  
\begin{equation}
\mathbf{v}^* = (\mathbf{X}^\top(\mathbf{X}\mathbf{X}^\top)^{-1}\mathbf{X}-\mathbf{I})\boldsymbol\beta_\Delta,
\end{equation}
where $\mathbf{X}\mathbf{X}^\top$ is assumed to be invertible. The derivation is included in the \gls{si}, Sec.\, \ref{SI:derivation_projection}. The expression can be simplified by inserting the singular value decomposition $\mathbf{X} = \mathbf{U\Sigma V}^\top$, 
\begin{equation}
    \mathbf{v}^* = (\mathbf{V} \mathbf{\Sigma}^\top (\mathbf{\Sigma} \mathbf{\Sigma}^\top)^{-1} \mathbf{\Sigma V}^\top -\mathbf{I})\boldsymbol\beta_\Delta,
    \label{eq:svd_closed1}
\end{equation}
which can be used to improve the numerical efficiency. Simplifying \eqref{eq:svd_closed1} leads to
\begin{equation}
    \mathbf{v}^* = \!\left(\!\mathbf{V} \begin{bmatrix}
  \mathbf{I}_n & \mathbf{0} \\
  \mathbf{0} & \mathbf{0} \end{bmatrix} \mathbf{V}^\top -\mathbf{I}\!\right)\!\boldsymbol\beta_\Delta.
\end{equation}
The property that $\mathbf{V}$ is an orthogonal matrix leads to
\begin{equation}
    \mathbf{v}^* = - \mathbf{V}\! \begin{bmatrix}
  \mathbf{0} & \mathbf{0} \\
  \mathbf{0} &  \mathbf{I}_{p-n} \end{bmatrix} \! \mathbf{V}^\top\boldsymbol\beta_\Delta.
  \label{eq:second_analytical_expression}
\end{equation}
The projection onto the nullspace can be a hard requirement that might yield a vector $\mathbf{v}^*$ that is dominated by noise and difficult to interpret, in particular, if $\mathbf{X}\mathbf{X}^{\top}$ is ill-conditioned as is often the case for many real-world chemical engineering problems.  
Furthermore, regularization shapes regression coefficients by trading their variance against a bias towards zero to improve generalization. However, regularized regression coefficients usually differ from the true coefficients (if they exist), and their difference is not expected to lie exactly within the nullspace but might be close to it, motivating the relaxed optimization.%
 
We propose to reformulate the optimization in \eqref{eq:opt_simple_orig} to allow deviations from the nullspace, 
\begin{equation}
\min_{\mathbf{v}} \  \normx[2]{
            \boldsymbol\beta_\Delta +\mathbf{v} }^2 + \gamma \normx[2]{\mathbf{Xv} }^2,
    \label{eq:min_diff_relaxed_constrains}
\end{equation}
where $\gamma$ is a nonnegative scalar.
Setting the derivative of  \eqref{eq:min_diff_relaxed_constrains} with respect to $\mathbf{v}$ to zero gives
\begin{equation}
    \mathbf{v}_\gamma   = -(\gamma \mathbf{X}^\top\mathbf{X}+\mathbf{I})^{-1}\boldsymbol\beta_\Delta.
\end{equation}
For $\gamma = 0$, the nullspace is not considered and  $\boldsymbol{\beta}_\Delta = \mathbf{v}_0$. For $\gamma \to \infty$, the optimization converges to \eqref{eq:second_analytical_expression}, as seen by
\begin{align}
    \lim_{\gamma \to \infty} -(\gamma \mathbf{X}^\top\mathbf{X}+\mathbf{I})^{-1}\boldsymbol\beta_\Delta
    &= \lim_{\gamma \to \infty} -\mathbf{V}(\gamma\boldsymbol{\Sigma}^{\top}\boldsymbol\Sigma + \mathbf{I})^{-1}\mathbf{V}^{\top}\boldsymbol\beta_\Delta \nonumber \\ 
    &= - \mathbf{V}\! \begin{bmatrix}
  \mathbf{0} & \mathbf{0} \\
  \mathbf{0} &  \mathbf{I}_{p-n} \end{bmatrix}\!  \mathbf{V}^\top\boldsymbol\beta_\Delta.
\end{align}
Analyzing the nullspace, i.e., comparing the coefficients $\boldsymbol{\beta}^\text{A}$ and $\boldsymbol{\beta}^\text{A} + \mathbf{v}_\gamma$ with $\boldsymbol\beta^\text{B}$, allows to identify which differences can be removed with a vector that is close to the nullspace and which differences would require significant deviations from the nullspace and are thus mainly responsible for the differences of the associated predictions. We propose to select $\gamma$, i.e., the penalization strength for deviations from the nullspace, based on a change in prediction accuracy to make it easier to interpret the result. That is, we define $\gamma$ based on the \gls{nrmse} defined by
\begin{align}
    s &= \max_i \{y_i\} - \min_i \{y_i\} \\
    L(\mathbf{\hat{y}},\mathbf{y}) &= \frac{1}{s\sqrt{n}}  \normx[2]{\mathbf{\hat{y}} - \mathbf{y}}, %
\end{align}
leading to the heuristic:
\begin{align}
&\!\max_{\gamma} \ \ \gamma 
\label{eq:optimization_gamma} \\
&\text{subject to\,}
\left| L(\mathbf{X}(\boldsymbol\beta^\text{A} + \mathbf{v}_\gamma), \mathbf{y}) - L(\mathbf{X}(\boldsymbol\beta^\text{A}), \mathbf{y}) \right| \! \leq c \notag \\
&\qquad\qquad\ \mathbf{v}_\gamma   = -(\gamma \mathbf{X}^\top\mathbf{X}+\mathbf{I})^{-1}\boldsymbol\beta_\Delta \notag
\end{align}
where $c$ defines the maximum loss function change introduced by the nullspace approach that is considered acceptable. The optimization %
\eqref{eq:optimization_gamma} is not convex for most practical examples but is easy to solve because it only has one degree of freedom, $\gamma$.\footnote{For example, the optimization can be solved by plotting the left-hand side of the inequality with respect to $\gamma$.}

%% file: content/04Examples.tex
\section{Case Studies}
\label{seq:examples}
This section demonstrates the nullspace method on several example cases to derive insights for interpretation of regression coefficients. The data $\mathbf{X}$ and $\mathbf{y}$ are generated synthetically for the first example. The second and third examples are on data from lithium-ion batteries \cite{severson2019data}, where we use constructed response variables by assuming different linear relationships to showcase the differences between regression coefficients and true coefficients. The last example uses the measured cycle life response where the true relationship between $\mathbf{X}$ and $\mathbf{y}$ is unknown.

\subsection{Synthetic Parabolic Data}
The parabolic example is inspired by measurements of some quantity over a continuous domain (e.g., time, concentration, voltage) to keep the data and relationships simple. The data are drawn from
\begin{align}
    \mathbf{x}_i &= a_i\mathbf{d}\odot\mathbf{d}, \quad i \in \{1,2,3, \cdots{}, 50\}, \\
    \mathbf{d} & = [1.0, 1.01, 1.02, \cdots{}, 3.0], \\
    \mathbf{X}^* &= \begin{bmatrix}
           \mathbf{x}_{1},
           \mathbf{x}_{2},
           \cdots
           \mathbf{x}_n
         \end{bmatrix}^\top,  \notag
\end{align}
where $\mathbf{d}$ is the vector of discretizations on the underlying domain, with a constant spacing of $0.01$ and a length of $p=201$, and $\odot$ is the element-wise product. The parameters $a_i \sim \mathcal{N}(\mu,\,\sigma^{2})$ with $\mu=0.3$ and $\sigma=0.3$. Consequently, $\mathbf{X} \in \mathbb{R}^{50 \times 201}$. We define the response as
\begin{align}
    \mathbf{y}^* &= \mathbf{X}^*\boldsymbol\beta^*, \notag \\ %
    \text{with} \ \boldsymbol\beta^* &= \frac{1}{p}\mathbf{I}.
    \label{eq:mean_i}
\end{align}
The true coefficients are thus equal. Subsequently, we add white Gaussian noise to the data and response
\begin{align}
    \mathbf{x}_i &= \mathbf{x}_i^* + \boldsymbol\epsilon_{xi} \\ 
    \mathbf{y} &= \mathbf{y}^* + \boldsymbol\epsilon_y
\end{align}
yielding the matrices $\mathbf{X}$ and $\mathbf{y}$ for use in regression. The added noise $\boldsymbol\epsilon_{xi} \in \mathbb{R}^{201}$ is chosen such that the average \gls{snr} of each sample  ($\mathbf{x}_i$) is 50 and the \gls{snr} of $\mathbf{y}$ is 50 as well. %
\begin{figure}[tbh]
\begin{center}
    \includegraphics[width=\hsize]{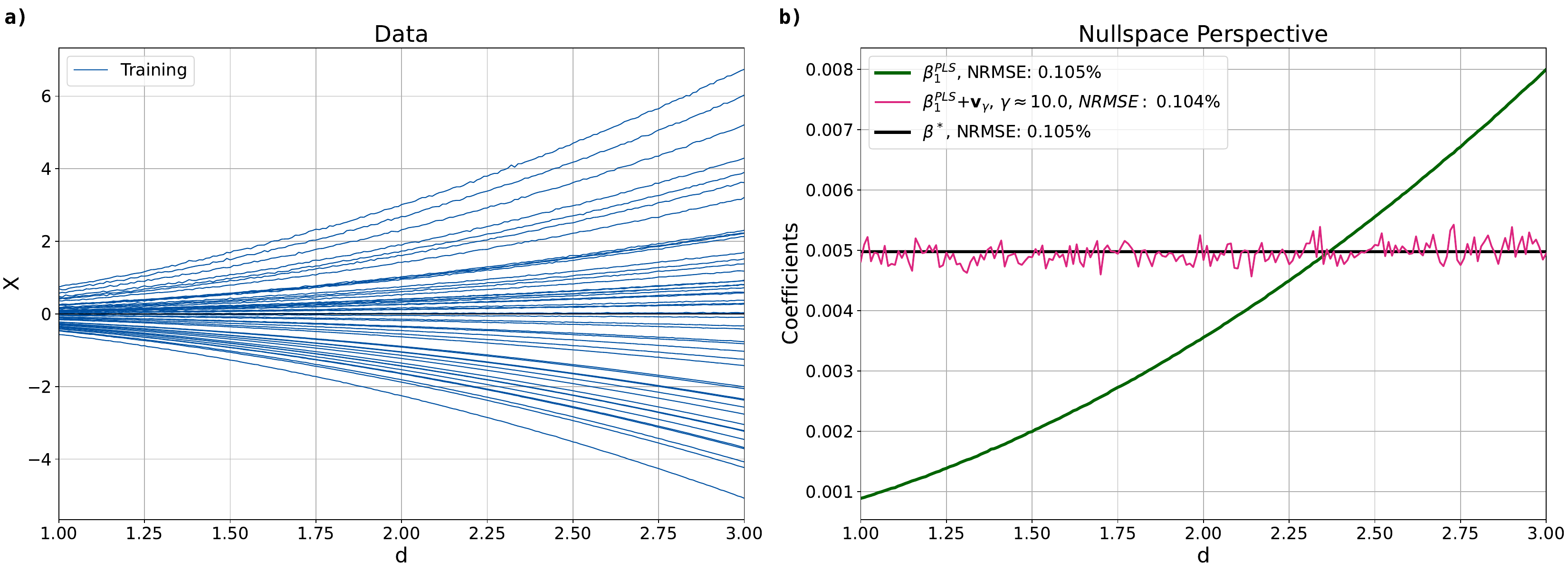}
\end{center}
    \caption{\textbf{a)} Mean centered parabolic data, with white Gaussian noise corresponding to an \gls{snr} of 50 added to $\mathbf{X}$ and $\mathbf{y}$ prior to mean centering, \textbf{b)} true coefficients in black and regression coefficients in green and nullspace-modified \gls{pls} coefficients in magenta.}
    \label{fig:parab_ex}
\end{figure}
Figure \ref{fig:parab_ex}a shows the mean-centered data, where each line corresponds to a matrix row. The 201 individual data points of each row are connected with a line, which is a reasonable visualization because of the underlying functional structure. We picked a \gls{pls} model with one component to learn the relationship between $\mathbf{X}$ and $\mathbf{y}$. The \gls{pls} method is popular among chemical engineers, and its regularization parameter, the number of components, is discrete and simple to choose. Figure \ref{fig:parab_ex}b  shows that the true coefficients and the \gls{pls} coefficients have very different shapes. However, their predictions and prediction accuracies are almost identical (cf.\ \gls{si}, Sec.\,\ref{si:parab_prediction_results}). The noise leads to a prediction error even when the true coefficients are used (i.e., 0.105\% \gls{nrmse}). Using the proposed nullspace method with hand-selected $\gamma=10$ to compare the true coefficients with the \gls{pls} coefficients shows that the resulting vector $\mathbf{v}_{10}$ is very close to the nullspace, i.e., does not significantly change the prediction accuracy and yields the adjusted coefficients $\boldsymbol\beta_1^{\text{PLS}} + \mathbf{v}_{10}$ that are very similar to the true coefficients $\boldsymbol\beta^*$. While $\boldsymbol\beta_1^{\text{PLS}}$ is orthogonal to the nullspace, $\boldsymbol\beta_1^{\text{PLS}} + \mathbf{v}_{10}$ is not orthogonal to the nullspace.
Due to the simple underlying structure of the data and the model, the \gls{pls} coefficients yield a similar prediction accuracy. However, the \gls{pls} coefficients have a smaller L$_2$-norm, i.e., $\normx[2]{\boldsymbol\beta^{\text{PLS}}_1}^2 < \normx[2]{\boldsymbol\beta^*}^2$, due to the implicit regularization of \gls{pls}. 

Assume that the coefficients are expected to be piecewise constant for physical reasons. We can then reformulate the regression as a generalized lasso problem with the matrix $\mathbf{D}$ in \eqref{eq:gen_lasso} set to $\mathbf{D}_1$.
\begin{figure}[tbh]
\begin{center}
    \includegraphics[trim={27.5cm 0 0 0}, clip, width=.6\hsize]{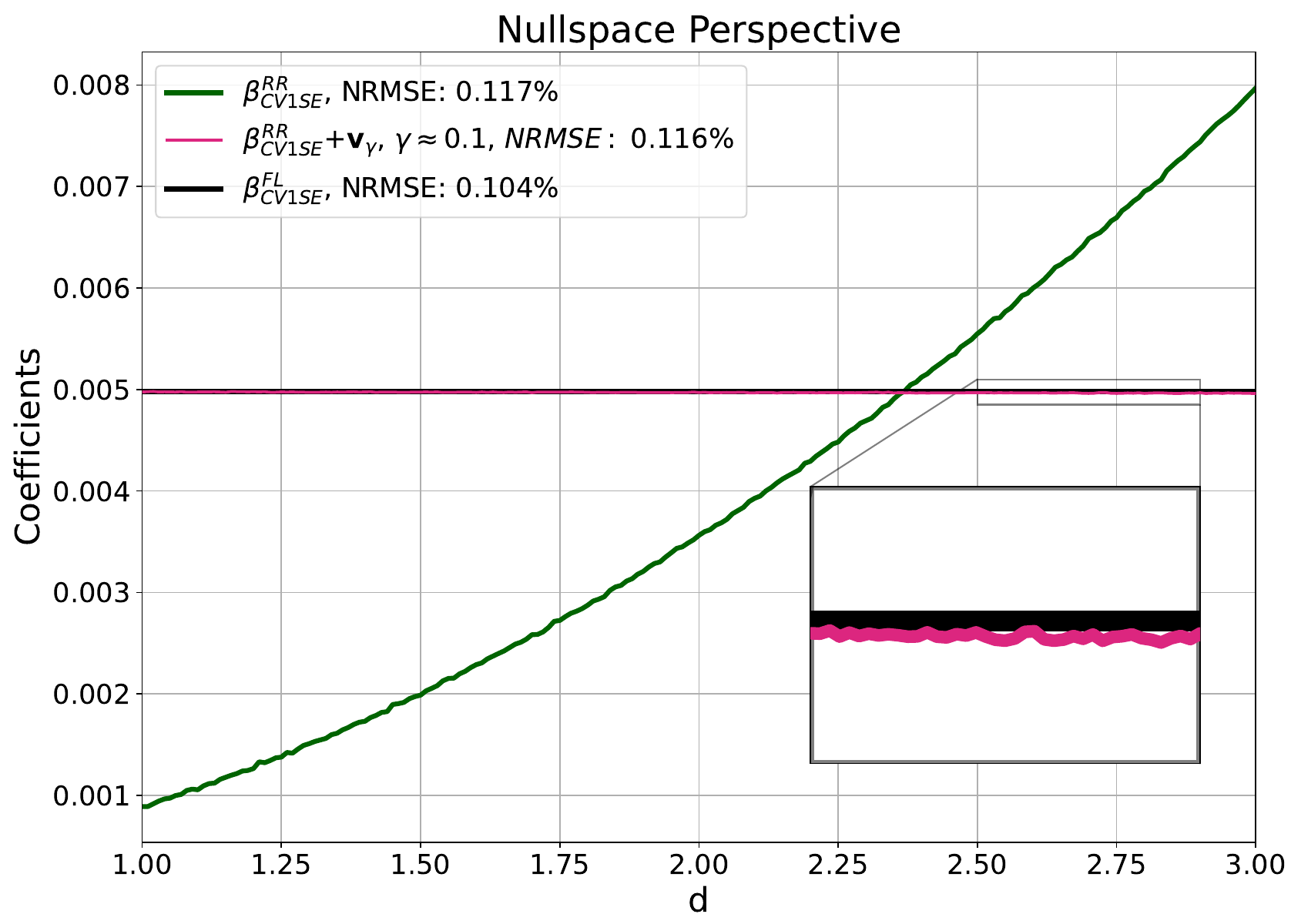}
\end{center}
    \caption{Ridge coefficients in green and fused lasso coefficients in black based on \glsentryshort{cv} and the one-standard-error rule. Nullspace-modified ridge coefficients in magenta.}
    \label{fig:parab_ex_fl_rr}
\end{figure}
Figure \ref{fig:parab_ex_fl_rr} shows the regression coefficients associated with ridge regression and the fused lasso. The regularization parameter was chosen by \gls{cv} and the one-standard-error rule \cite{hastie2009elements}. Figure \ref{fig:parab_ex_fl_rr} looks remarkably similar to Fig.\ \ref{fig:parab_ex}. The fused lasso coefficients are nearly identical to the true coefficients, and the ridge coefficients are similar to the \gls{pls} coefficients with one component but slightly noisier. 

From the data alone, it is not possible to state whether $\mathbf{y}$ was constructed from constant or parabolic coefficients. Furthermore, regression coefficients obtained from methods that are orthogonal to the nullspace can yield coefficients that appear to disagree with prior knowledge at first sight. As this example shows, methods that are not orthogonal to the nullspace such as the fused lasso can be advantageous for interpretation and conclusions if selected based on prior knowledge.

\subsection{Lithium-Ion Battery Data}
As a real-world measurement data example, we consider a \gls{lfp} battery dataset, which contains cycling data for 124 batteries \cite{severson2019data}. Each battery has a fixed charging and discharging protocol. The charging protocols vary widely between the cells, whereas the discharge is constant and identical for all cells. The objective of the original paper was the prediction of the cycle life, i.e., the number of cycles until the battery's capacity drops below $80\%$ of its nominal capacity. Features based on the difference between the discharge capacity of voltage curves for two cycles, subsequently called $\Delta \mathbf{Q}_{\text{a}-\text{b}}$, were shown to linearly correlate well with the logarithm of the cycle life. For this case study, we use the cycle pair $a=100$ and $b=10$, as done in \cite{severson2019data}. Furthermore, we denote by a tilde ($\Delta \widetilde{\mathbf{Q}}_{100-10}$) that the columns are mean centered. The dimensionality $\Delta\mathbf{Q}_{100-10} \in \mathbb{R}^{41 \times 1000}$ due to the high resolution of the discharge capacity over the voltage domain. More information about the data set and reasoning about the modeling objective can be found in \cite{severson2019data}. 
\begin{figure}[tbh]
\begin{center}
    \includegraphics[width=\hsize]{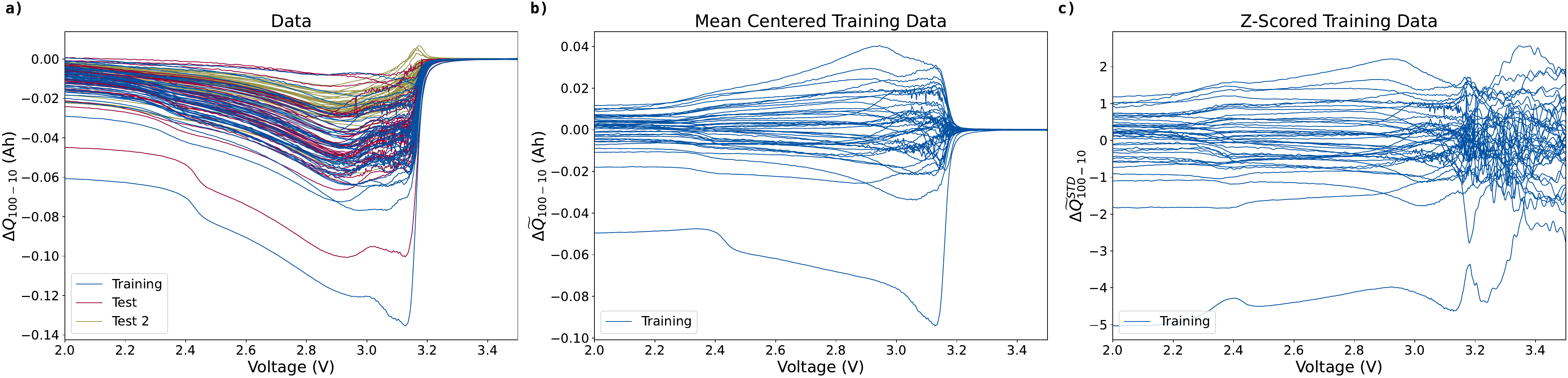}
\end{center}
    \caption{\textbf{a)} \gls{lfp} Discharge capacity difference between cycle 100 and 10, data split into training, primary and secondary test set, \textbf{b)} mean centered training data, \textbf{c)} z-scored training data.}
    \label{fig:data_lfp} 
\end{figure}
Figure \ref{fig:data_lfp}a shows the \gls{lfp} data set, partitioned into training, primary, and secondary test data as suggested in \cite{severson2019data}. Figure \ref{fig:data_lfp}b shows the mean subtracted training data. The data of the shortest-lived battery is clearly separated from the remainder of the data set. However, we keep this battery in the data set, as its influence on the training is benign. Figure \ref{fig:data_lfp}c shows the z-scored training data (i.e., standardized data, yielding unit variance columns). The unit (Ah) is lost by z-scoring the data. Usually, z-scoring is recommended for data with features that have different units and thus might vary by orders of magnitude. However, for functional high-dimensional data, the unit of all columns is the same. Nevertheless, the measured values can vary by order of magnitude. Figure \ref{fig:data_lfp}c shows that the noise in the voltage region 3.2--3.5\,V is amplified by rescaling because of a lower signal-to-noise ratio in this voltage region. A more detailed analysis of the \gls{snr} can be found in the \gls{si}, Sec.\,\ref{si:snr_lfp}. However, whether z-scoring is useful does not only depend on the data matrix $\mathbf{X}$  but also on its underlying relationship with $\mathbf{y}$, which we explore next on synthetic responses $\mathbf{y}$ before moving to the cycle life response.

\subsubsection{Synthetic Response}
\paragraph{Constant Coefficients.} The response for this example is the sample mean defined in \eqref{eq:mean_i}, with $p=1000$ to match the dimensionality of the \gls{lfp} data set with added white Gaussian noise corresponding to an \gls{snr} of 50.  
\begin{figure}[htb]
    \centering
    \begin{subfigure}{.495\linewidth}
      \centering
      \includegraphics[trim={28cm 0 0 0}, clip, width=.98\linewidth]{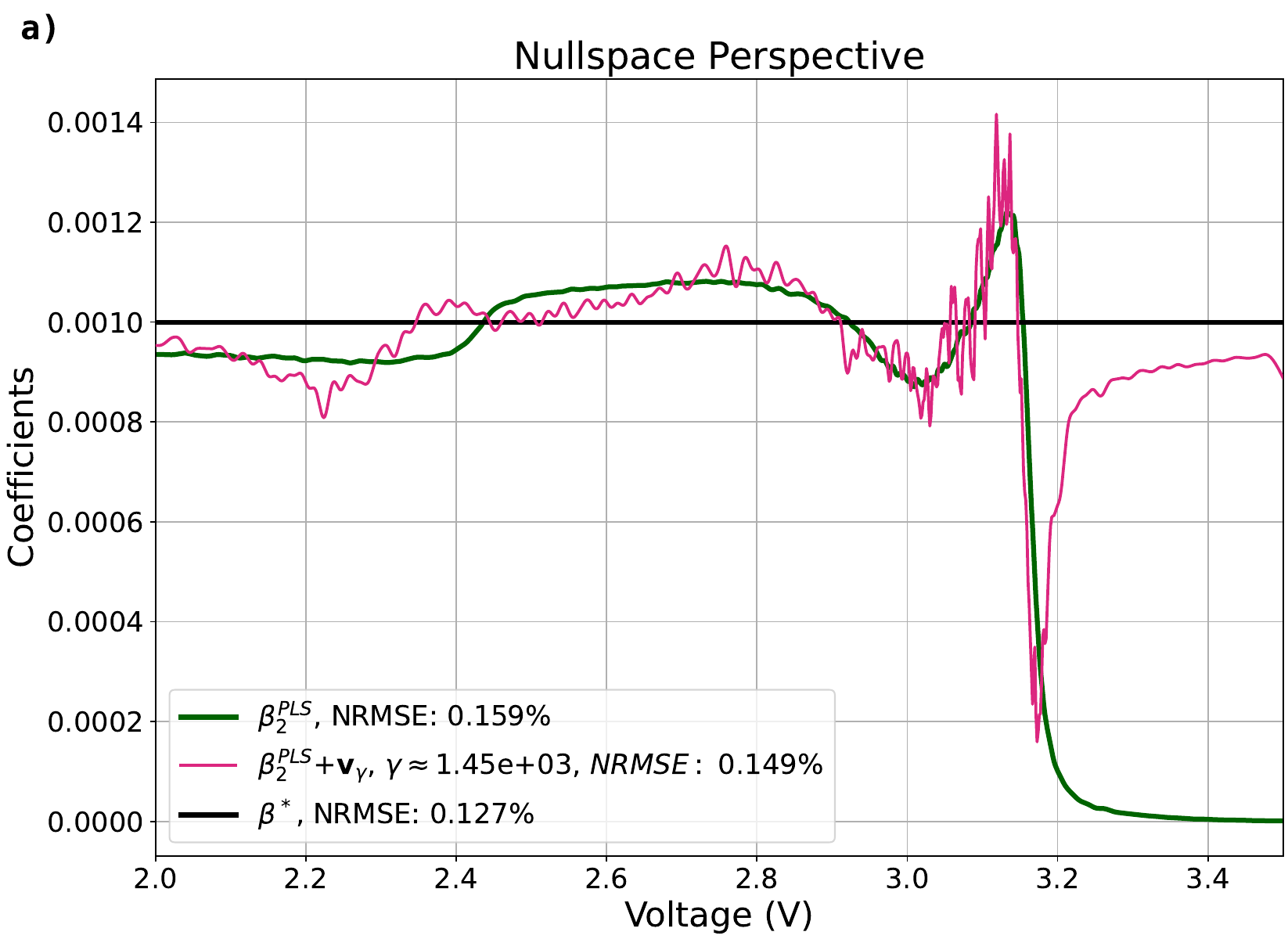}
      \label{fig:LFP_mean_2PLS}
    \end{subfigure}
    \begin{subfigure}{.495\linewidth}
      \centering
      \includegraphics[trim={27.5cm 0 0 0}, clip, width=.98\linewidth]{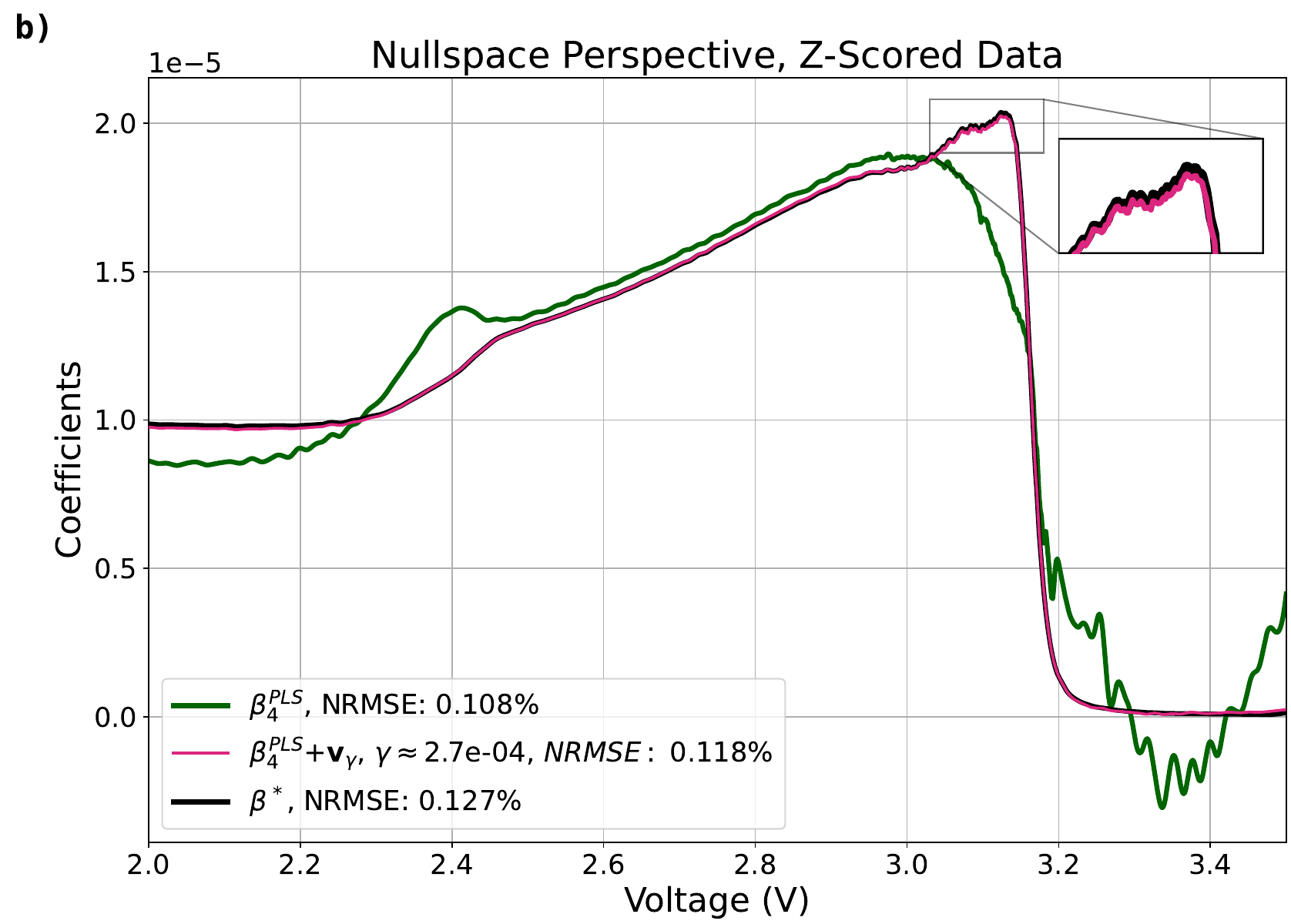}
      \label{fig:LFP_mean_std_4PLS}
    \end{subfigure}
    \caption{\textbf{a)} True coefficients in black, \gls{pls} coefficients based on \gls{cv} and the one-standard-error deviation rule in green, nullspace-modified coefficients in magenta, \textbf{b)} nullspace perspective similar to a, with \gls{pls} coefficients estimated with the one-standard-error rule corresponding to z-scored data.}
    \label{fig:LFP_mean}
\end{figure}
Figure \ref{fig:LFP_mean}a shows the nullspace perspective for the constant coefficient response with the data on the original scale, and Fig.\,\ref{fig:LFP_mean}b based on z-scored data (i.e., columns of $\mathbf{X}$ are scaled to have a unit standard deviation). The number of \gls{pls} components is determined by \gls{cv} and the one-standard-error rule. The \gls{pls} model associated with the z-scored data needs more components. The nullspace penalization parameter $\gamma$ was chosen in both cases to yield $c=0.01\%$ \gls{nrmse} prediction error change. Figure \ref{fig:LFP_mean}a shows the differences between the true coefficients and the \gls{pls} model's regression coefficients in the section from 2.0--3.1\,V are relatively close to one another; however, some differences remain. The differences in the voltage region from 3.2 to 3.5\,V only have a minor effect on the prediction results. Most of the difference between the regression coefficients in this area is associated with the nullspace, indicated by the large difference between the nullspace-modified \gls{pls} coefficients in magenta and the original \gls{pls} coefficients in green. Thus, the differences in the region 3.2 to 3.5\,V do not change the prediction results on the training data significantly and arise due to the interplay of the regularization objective with the nullspace.

When the data are z-scored, most of the visible differences between the \gls{pls} coefficients in green and the true coefficients in black are contained in the enlarged nullspace (Fig.\ \ref{fig:LFP_mean}b). The modified coefficients match the true coefficients very well. The prediction error difference between the \gls{pls} coefficients in green and the nullspace-modified coefficients in magenta is 0.01\% \gls{nrmse}. The remaining differences between the true coefficients in black and the modified coefficient in magenta are barely visible but are responsible for another 0.01\% \gls{nrmse} prediction error change, highlighting the effect of the nullspace. Comparing Figs.\ \ref{fig:LFP_mean}a and \ref{fig:LFP_mean}b shows that, in case the true coefficients are constant (i.e., all columns are equally important), z-scoring can help regression to yield coefficients that are more similar to the true coefficients. 

\begin{figure}[htb]
\centering
\begin{subfigure}{.495\linewidth}
  \centering
  \includegraphics[trim={28cm 0 0 0}, clip, width=.98\linewidth]{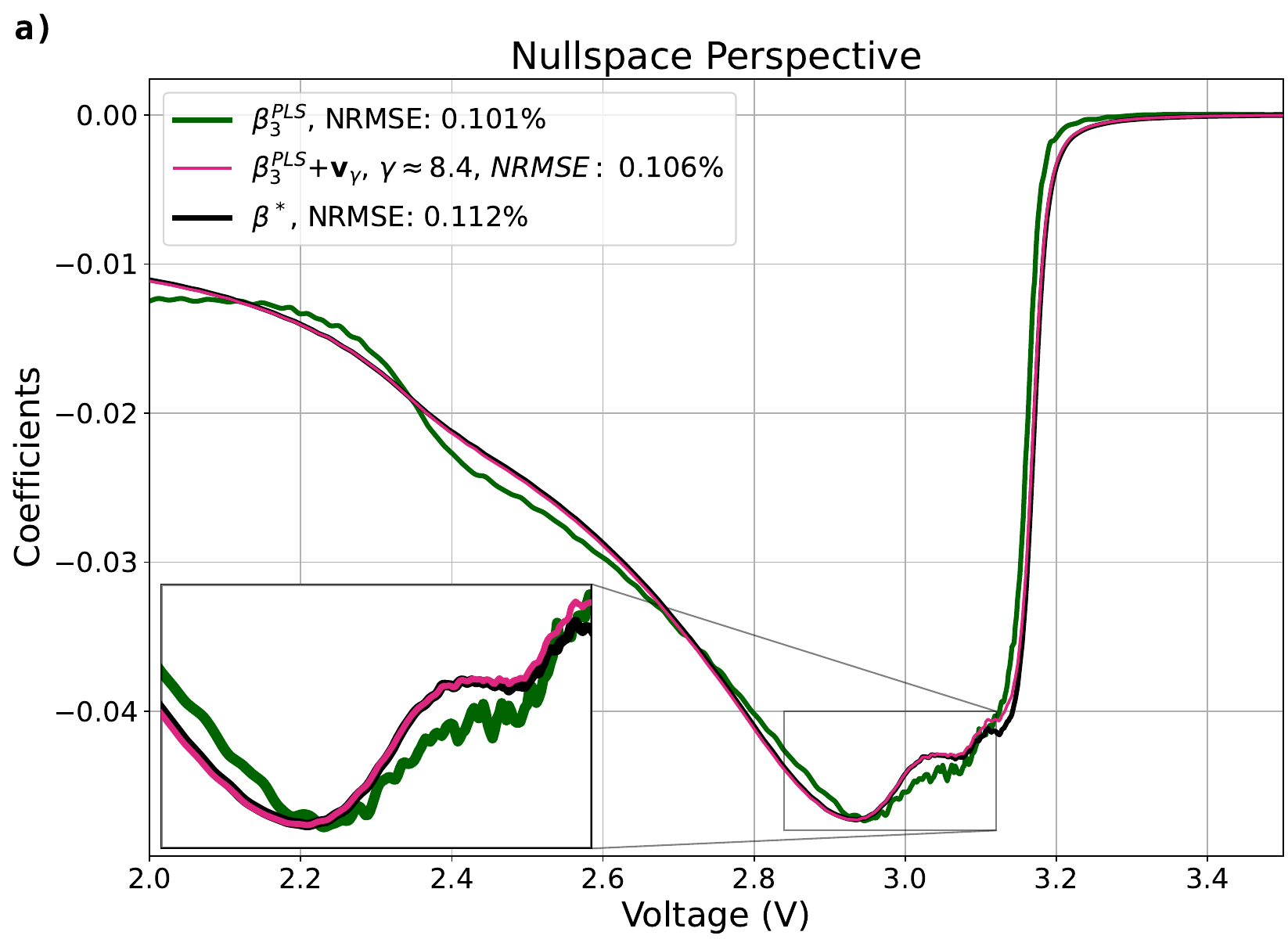}
  \label{fig:LFP_cm_3pls}
\end{subfigure}
\begin{subfigure}{.495\linewidth}
  \centering
  \includegraphics[trim={27.5cm 0 0 0}, clip, width=.98\linewidth]{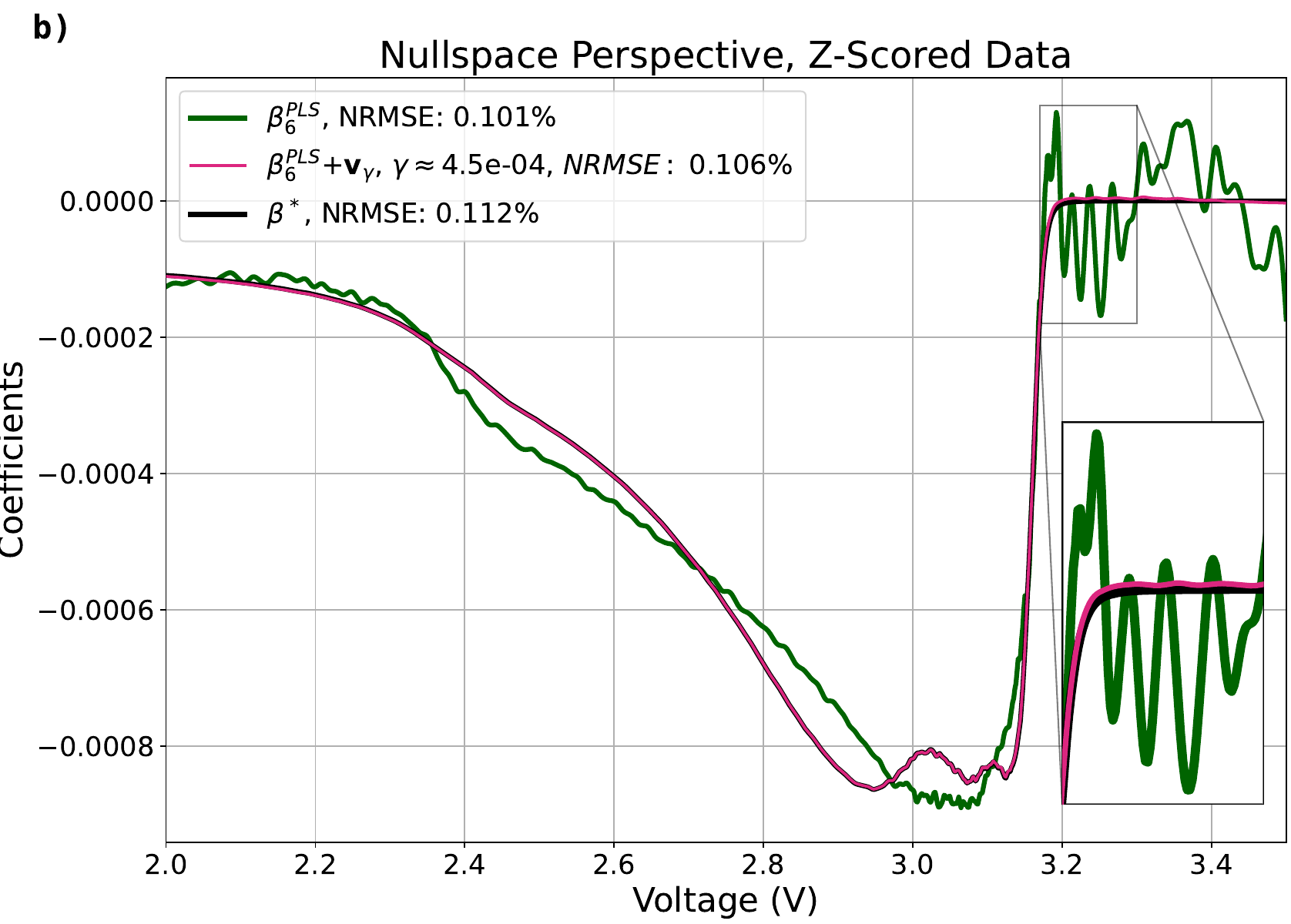}
  \label{fig:LFP_cm_std_6PLS}
\end{subfigure}
\caption{\textbf{a)} True coefficients in black, \gls{pls} coefficients based on \gls{cv} and the one-standard-error rule in green, nullspace-modified coefficients in magenta, \textbf{b)} nullspace perspective similar to a, with \gls{pls} coefficients estimated with the one-standard-error rule corresponding to z-scored data.}
\label{fig:LFP_cm}
\end{figure}

\paragraph{Column Mean Coefficients.}
The true coefficient vector $\boldsymbol\beta^*$ for the next synthetic example is the column mean of the data $\mathbf{X}$ prior to column centering
\begin{align}
    \beta_j^* &= \frac{1}{n}\sum_{i=1}^n x_{i,j}, \\
    \boldsymbol\beta^* &= \!\begin{bmatrix}
           \beta_{1}^*,
           \beta_{2}^*,
           \cdots{},
           \beta_{p}^*
         \end{bmatrix}^\top\!.\notag
\end{align}
The \gls{pls} model with 6 components associated with the z-scored data picks up a high amount of noise in the voltage regions from 3.3 to 3.5\,V (Fig.\ \ref{fig:LFP_cm}b). In contrast, the \gls{pls} model with 3 components associated with the data on the original scale converges well to the true coefficients over the entire voltage region (Fig.\ \ref{fig:LFP_cm}a). The small differences are very closely associated with the nullspace. Here, z-scoring amplifies and feeds noise into the model, manifesting as the spiky regression coefficients, with the most extreme spikes in the voltage regions from 3.2 to 3.5\,V (Fig.\ \ref{fig:LFP_cm}b). Still, the \gls{pls} model associated with the z-scored data has approximately the same prediction accuracy as the \gls{pls} model associated with the original data.

Suppose there was some prior evidence or physical intuition that the true coefficients are constant or at least of similar magnitude. Then, z-scoring feeds the assumption that all the columns' importance is in the same order of magnitude to the model. However, if the coefficients are expected to vary by an order of magnitude (e.g., as is the case for the true coefficients being the column mean of the data), then not z-scoring the data accounts for the assumption that the scale of the columns is correlated with the assumed underlying true coefficients. 
The two examples %
show that the potential effects of z-scoring on the regression coefficients should be considered carefully for functional data. When data are z-scored, the model can become better at learning the underlying relationship, but noise may be amplified, depending on the noise structure.

\subsubsection{Measured Cycle Life Response}
The measured response associated with the \gls{lfp} battery data is the cycle life. We train the models by using the logarithm of the cycle life and use the same training, primary test, and secondary test set as suggested in \cite{severson2019data}. %
We determine the regularization parameter based on the minimum \gls{cv} error and do not employ the one-standard-error rule.\footnote{The standard deviation of the \gls{cv} error is large due to the long-living cells that heavily influence the prediction performance, which would lead to overly conservative regularization estimates.}
\begin{figure}[tbh]
\begin{center}
    \includegraphics[width=\hsize]{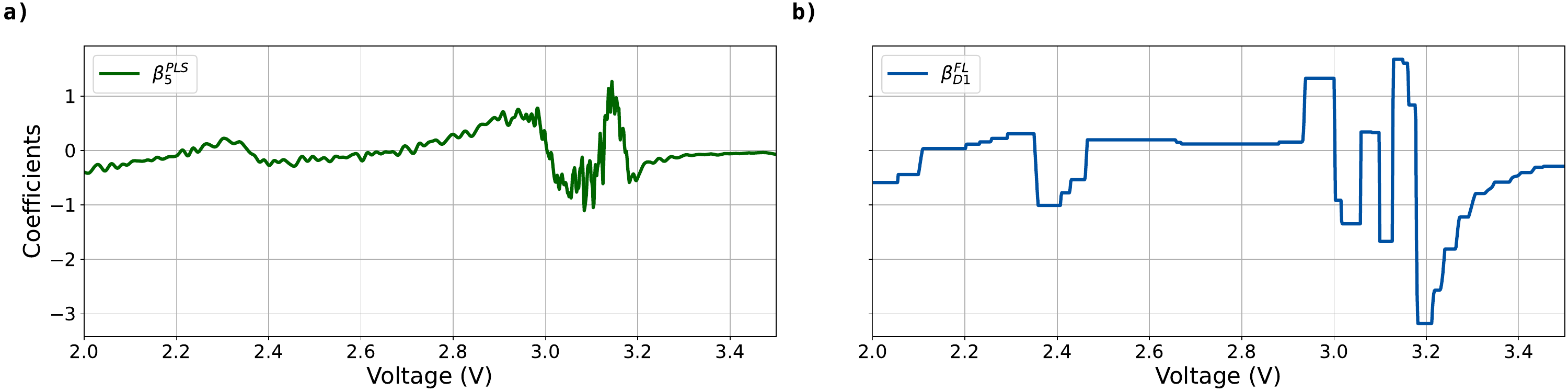}
\end{center}
    \caption{Cross-validated regression coefficients, original data (cf.\ Fig.\ \ref{fig:data_lfp}b): \textbf{a)} \gls{pls} coefficients, \textbf{b)} fused lasso coefficients.}
    \label{fig:reg_coef_gen_lasso}
\end{figure}
The \gls{pls} coefficients with five components have a similar shape as the fused lasso coefficients 
(cf.\ Figs.\,\ref{fig:reg_coef_gen_lasso}ab). However, the \gls{pls} coefficients have high-frequency perturbations, in particular, in the voltage range from 2.9--3.2\,V, which is likely due to noise, making the \gls{pls} coefficients harder to interpret. 
The fused lasso coefficients (Fig.\ \ref{fig:reg_coef_gen_lasso}b) clearly indicate three regions of importance, enabling a physical interpretation. The range around 2.0--2.1\,V is associated with the capacity change of the cell between cycles 10 and 100. Around 2.4\,V, a different pattern can be seen in the data (Fig.\ \ref{fig:data_lfp}), corresponding to the negative peak in the regression coefficients, which may correspond to \gls{lfp} cathode degradation associated with iron anti-site defects, as the free energy of reaction (overpotential times charge) exceeds their formation energy ${\sim}0.55$ eV \cite{rahul_particle_size2011}. This interpretation is also consistent with experiments showing that chemical reduction of \gls{lfp} by citric acid is able to heal iron anti-site defects with a similar free energy of reaction of 0.58 eV
\cite{xu_meng2020}.
The voltage range around 2.9--3.3\,V contains most of the regression coefficient peaks. The two dominant plateaus of the \gls{ocv}, which result from the single broad plateau of \gls{lfp} superimposed with two more narrow plateaus of graphite, are located here, and most of the cell’s capacity is discharged in this voltage range. These voltage plateaus correspond to phase transformations of the porous electrodes \cite{Ferguson_2012_nonequ_thermodymn_por_elec}, specifically between the low and high-density stable phases of \gls{lfp}, as well as between stages 1, 2, and 3 of lithiated graphite \cite{Ferguson_2014_phase_transformation_dynamics}. The fused-lasso coefficients showcase three distinct negative and positive peaks, corresponding to changes in the rate-dependent tilt of the voltage plateaus, which may result from changes in particle-size-dependent nucleation barriers and population dynamics of reaction-controlled phase transformations \cite{Ferguson_2014_phase_transformation_dynamics, li_2014_current_induces_transition}. The peak width and height can be interpreted as a weighted sum of the average slopes of the data between the respective peaks. On low-rate data, the position and magnitude of peaks in the incremental capacity analysis correspond to different degradation modes \cite{dubarry_synthesize_batt_degradation}. The peaks and peak shifts of the incremental capacity analysis blur out at higher C-rates, as expected from the suppression of phase separation by driven auto-inhibitory reactions \cite{bazant2017phase_sep_autocatalysis}. In particular, the decreasing reaction rate with increasing lithium concentration in the \gls{lfp} cathode, which has been predicted theoretically \cite{fraggedakis_2021_theory_couple_ion_electron_kin} and confirmed experimentally \cite{Zhao_2022_learning_pixel_by_pixel}, erases the voltage plateaus at high rates and causes more homogeneous reactions that are likely favorable for battery lifetime \cite{Ferguson_2012_nonequ_thermodymn_por_elec, Ferguson_2014_phase_transformation_dynamics, li_2014_current_induces_transition}. However, the obtained regression coefficients indicate that there is degradation information in this region even in $\boldsymbol\Delta\mathbf{Q}_{100-10}$ (i.e., the discharge capacity difference of cycle 100 and 10, both at 4C) that is important for capturing past degradation and forecasting future degradation. On the other hand, if the 4C current is well into the regime of suppressed phase separation, then we would expect a negative correlation between lifetime and internal resistance of the intercalation reaction, which in turn is correlated with larger $\boldsymbol\Delta\mathbf{Q}_{100-10}$.

\begin{figure}[tbh]
\begin{center}
    \includegraphics[width=\hsize]{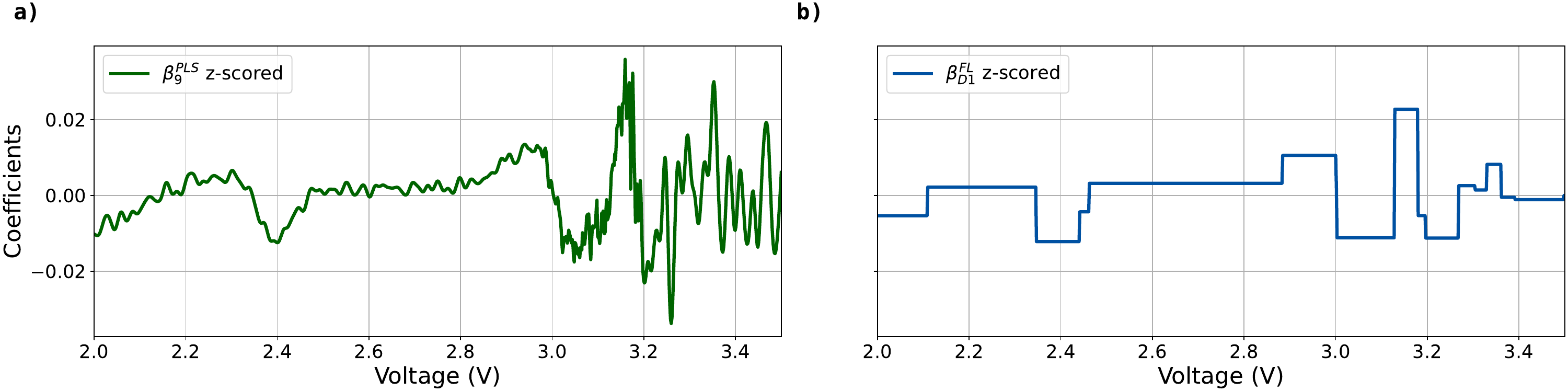}
\end{center}
    \caption{Cross-validated regression coefficients, z-scored data (cf.\ Fig.\ \ref{fig:data_lfp}c): \textbf{a)} \gls{pls} coefficients, \textbf{b)} fused lasso coefficients.}
\label{fig:reg_coef_gen_lasso_std}
\end{figure}
The coefficients regressed on the z-scored data have similar peaks and characteristics as the coefficients regressed on the original data (cf.\ Figs.\ \ref{fig:reg_coef_gen_lasso}ab, and \ref{fig:reg_coef_gen_lasso_std}ab). 
The z-scoring of columns introduces a linear transformation that significantly changes the regression coefficients in the range from 3.2 and 3.4\,V. 
\begin{table}[htb]
  \caption{\gls{rmse} prediction accuracies associated with the coefficients in Figs.\ \ref{fig:reg_coef_gen_lasso}--\ref{fig:reg_coef_gen_lasso_std}. Low \gls{cl}: $y_i \leq 1200$ cycles; high \gls{cl}: $y_i > 1200$ cycles. All models were trained on the entire training data.}
  \label{tab:pred_res_cl_overview}
  \centering
  \begin{tabular}{l|cc|cc|c}
    \toprule
     & \multicolumn{2}{c|}{Original Scale} & \multicolumn{2}{c|}{Z-Scored} & Feature\\
    \midrule
    Set & FL D1 & PLS 5 Comp.\ & FL D1 & PLS 9 Comp.\ \cite{attia_learning_lfp} & Variance Model \cite{severson2019data}\\
    \midrule
    \textit{Training (41)}           &  \textit{68} &           \textit{83} &           \textit{62} &                    \textbf{\textit{57}} &             \textit{104} \\
    Test 1 (42)             & 115 &          116 &          105 &                   \textbf{102} &             138 \\
    Test 2 (40)             & 198 &          217 &          192 &                   \textbf{174} &             196 \\
    \midrule
    \midrule
    \textit{Training Low CL (39)}    &  \textit{62} &          \textit{ 82} &          \textit{ 53} &                    \textbf{\textit{50}} &             \textit{103} \\
    Test 1 Low CL (39)      &  96 &          101 &           \textbf{76} &                    80 &              96 \\
    Test 2 Low CL (34)      & 135 &          202 &          \textbf{115} &                   132 &             119 \\
    \midrule
    \textit{Training High CL (2)}    & \textit{138} &          \textbf{\textit{106}} &          \textit{150} &                  \textit{ 139} &             \textit{115} \\
    Test 1 High CL (3)      & 258 &          \textbf{231} &          280 &                   252 &             385 \\
    Test 2 High CL (6)      & 395 &          \textbf{285} &          412 &                   322 &             419 \\
    \bottomrule
  \end{tabular}

\end{table}

The fused lasso based on the z-scored columns yields high prediction accuracy and interpretable coefficients (cf.\ Tab.\ \ref{tab:pred_res_cl_overview} and Fig.\ \ref{fig:reg_coef_gen_lasso_std}b). In the higher voltage region, an additional peak appears around 3.35\,V, %
which could not be learned from the original data because of the very small variance of the data prior to rescaling in combination with regularization. 
Moreover, the coefficients estimated on the z-scored data have the highest prediction performance on the training, primary, and secondary test sets (Tab.\ \ref{tab:pred_res_cl_overview}), showcasing that there is valuable information in the higher voltage region above 3.2\,V.
Furthermore, both models on the z-scored data outperform the variance model suggested in \cite{severson2019data}. The \gls{pls} model with nine components, suggested first in \cite{attia_learning_lfp}, slightly outperforms the fused lasso model when all cells are considered. However, the fused lasso yields the lowest \gls{rmse} error for both test sets when only evaluated on the shorter-lived cells. The higher performance of the \gls{pls} model with nine components on the secondary test set is thus mainly associated with the longest-living cells that are more difficult to predict (cf.\  \cite{severson2019data, attia_learning_lfp}). But, the coefficients associated with the \gls{pls} model are challenging to interpret because their sign changes frequently. What is more, the secondary test set was impacted by a longer calendar aging due to an extended storing period before the cycling started (cf.\ \gls{si} of \cite{attia2020closed}), making it tough to understand the higher prediction accuracy of the \gls{pls} model on the secondary test set. In particular, the \gls{pls} coefficients show further peaks in the voltage region above 3.4\,V influenced at least partially by noise because the \gls{snr} in this region is very low (cf.\  \gls{si} Sec.\ \ref{si:snr_lfp}).

Similarly to the parabolic data set, we observe that the fused lasso coefficients are more interpretable than the \gls{pls} coefficients.
\begin{figure}[!htb]
\begin{center}
    \includegraphics[width=\hsize]{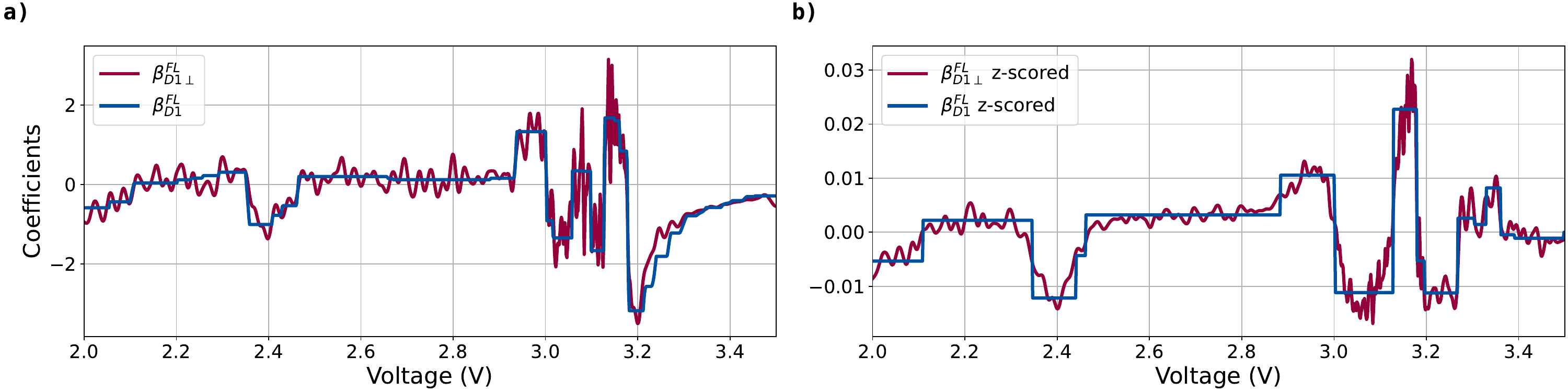}
\end{center}
    \caption{Comparison of fused lasso coefficients (blue) and their component orthogonal to the nullspace (red): \textbf{a)} original data, \textbf{b)} z-scored data.}
\label{fig:fl_orthogonal_comparison}
\end{figure}
Not requiring the coefficients orthogonal to the nullspace improves interpretability (Fig.\ \ref{fig:fl_orthogonal_comparison}ab). The component of the regression coefficients orthogonal to the nullspace (cf.\ red coefficients in Fig.\ \ref{fig:fl_orthogonal_comparison}ab) are less interpretable while making identical predictions on the training data.

%% file: content/05Conclusion.tex
\section{Conclusion}
\label{sec:conclusion}
The article proposes a nullspace perspective for gaining insights to help make informed design choices for building regression models on high-dimensional data and for reasoning about potential underlying linear models. 
We demonstrate the nullspace method on a fully synthetic dataset and lithium-ion battery data with a designed linear response. %
Applying the nullspace insights for predicting the cycle life led to further insights into how degradation manifests itself for \gls{lfp} batteries during discharge at 4C.

The nullspace allows different-looking regression coefficients to yield similar predictions (Fig.\ \ref{fig:parab_ex}). While z-scoring for high-dimensional functional data can be beneficial, it should be an active design choice because it can increase noise by scaling up columns with low \gls{snr} (Fig.\ \ref{fig:LFP_cm}). Appropriate regularization can mitigate increased noise after z-scoring. Furthermore, regularization and z-scoring must be carefully considered and correspond to prior physical knowledge to obtain interpretable regression results (Figs.\ \ref{fig:reg_coef_gen_lasso}, \ref{fig:reg_coef_gen_lasso_std}). Otherwise, the combination of the nullspace and regularization can hinder interpretability and potentially make it impossible to obtain regression coefficients close to the true coefficients. 

Regression methods which yield coefficients orthogonal to the nullspace, such as \gls{rr}, \gls{pcr}, and \gls{pls}, can be challenging to interpret. Methods that yield regression coefficients not orthogonal to the nullspace, such as the fused lasso, can be advantageous for interpretability (Fig.\,\ref{fig:fl_orthogonal_comparison}). 

The learnings from the nullspace analysis help to build and interpret linear regression models for high-dimensional functional data. The case studies show how to reason about underlying linear relationships between $\mathbf{X}$ and $\mathbf{y}$, which is important for system optimization and to improve scientific understanding.

%% file: content/SupplementaryInformation.tex
\clearpage
\makeatletter
\setcounter{equation}{0}
\setcounter{page}{1}
\setcounter{figure}{0}
\counterwithin{table}{subsection}
\renewcommand\thesection{S\@arabic\c@section}
\renewcommand\thetable{S\@arabic\c@table}
\renewcommand\thefigure{S\@arabic\c@figure}
\renewcommand\thepage{S\arabic{page}}
\renewcommand{\theequation}{S.\arabic{equation}}
\renewcommand{\thesubsection}{\Alph{subsection}}
\makeatother

\begin{refsection}

\section*{Supplementary Information for ``\titlecopy''}
\label{sec:si:sup_info}

\section{Ordinary Least Squares and Minimum Norm Solution}
\label{SI:derivation}
Ordinary Least Squares Regression finds a solution $\boldsymbol{\beta}^{\text{OLS}}$ that minimizes the L$_2$-norm of the regression error $\boldsymbol{\hat\epsilon}$ \cite{LA_GStrang},
\begin{equation}
    \min_{\boldsymbol{\beta}}  \|\mathbf{y}-\mathbf{X}\boldsymbol{\beta}\|_2^2.
    \label{eq:ols_obj}
\end{equation}
If $p < n$ and $\textrm{rank}(\mathbf{X}) = p$, it follows that $\mathbf{X^{\top}X}$ is invertible, and \eqref{eq:ols_obj} can be solved analytically to give
\begin{equation}
    \boldsymbol{\beta}^{\text{OLS}} = (\mathbf{X^{\top}X})^{-1}\mathbf{X^{\top}y}.
    \label{eq:OLSclosedform}
\end{equation}
\gls{ols} estimates have low bias and are optimal under the assumption of the Gauss-Markov theorem. However, they have a very large variance if the condition number for inversion of $\mathbf{X}^\top\mathbf{X}$ is large, as it the case for many real-world data analytics problems, resulting in low prediction accuracy on unseen data.

For $p < n$ and $\lambda \to 0$, \gls{rr} converges to \gls{ols}. In the more general case, without making assumptions about the dimensionality and rank of the real matrix $\mathbf{X}$, the \gls{svd}, $\mathbf{X} = \mathbf{U}\boldsymbol{\Sigma}\mathbf{V}^{\top}$, can be used to show that (e.g.,  \cite{kobak2020optimal}) 
\begin{align}
    \boldsymbol{\beta}_0 = \lim_{\lambda \to 0} \boldsymbol{\beta}_{\lambda} &= \lim_{\lambda \to 0} (\mathbf{X^{\top}X + \lambda \mathbf{I}})^{-1}\mathbf{X^{\top}y} \nonumber\\
    &= \lim_{\lambda \to 0} (\mathbf{V}\boldsymbol{\Sigma}^{\top}\boldsymbol\Sigma\mathbf{V}^{\top} + \lambda\mathbf{V}\mathbf{V}^{\top})^{-1}\mathbf{V}\boldsymbol{\Sigma}^{\top}\mathbf{U}^{\top}\mathbf{y} \nonumber\\
    &= \mathbf{V}\boldsymbol{\Sigma}^{\dagger}\mathbf{U}^{\top}\mathbf{y}\nonumber\\
    &= \mathbf{X}^{\dagger}y.
\end{align}

\section{Orthogonality of Coefficient Vector and Predictor Nullspace for RR, PCR, and PLS}
\label{SI:orthogonality}
This section shows the orthogonality of regression coefficients and the nullspace for \gls{rr}, \gls{pcr}, and \gls{pls}. Please note that the notation differs slightly from the main section to improve readability. 
To show the orthogonality for \gls{pcr} and \gls{pls}, the SVD of $\mathbf{X}$ is written in the partitioned form
\begin{align}
  \mathbf{X}
    & = \begin{bmatrix} \mathbf{U}_1 & \mathbf{U}_0 \end{bmatrix} \begin{bmatrix} \boldsymbol{\Sigma}_1 & \mathbf{0} \\ \mathbf{0} & \mathbf{0} \end{bmatrix} \begin{bmatrix} \mathbf{V}_1^{\top} \\[1mm] \mathbf{V}_0^{\top} \end{bmatrix}
    = \mathbf{U}_1 \boldsymbol{\Sigma}_1 \mathbf{V}_1^{\top}
\end{align}
where $\boldsymbol{\Sigma}_1$ contains only the non-zero singular values.
Thus the columns of $\mathbf{V}_0$ give an orthonormal basis of $\mathcal{N}(\mathbf{X})$.
As $\mathbf{V}$ is an orthogonal matrix, $\mathbf{V}_0^{\top} \mathbf{v}_i = \mathbf{0}$
holds for every column $\mathbf{v}_i$ of $\mathbf{V}_1$, or simply $\mathbf{V}_0^{\top} \mathbf{V}_1 = \mathbf{0}$.
The statement that a vector $\boldsymbol{\beta}$ is orthogonal to $\mathcal{N}(\mathbf{X})$ is equivalent to $\mathbf{V}_0^{\top} \boldsymbol{\beta} = \mathbf{0}$.

\paragraph{Ridge regression}
Ridge regression solves
\begin{align}
  \arg \min_{\hat{\boldsymbol{\beta}}} \|\mathbf{y} - \mathbf{X} \hat{\boldsymbol{\beta}}\|_2^2 + \gamma \| \hat{\boldsymbol{\beta}} \|_2^2\;.
\end{align}
Any $\boldsymbol{\beta}$ can be written as $\boldsymbol{\beta} = \boldsymbol{\beta}_1 + \boldsymbol{\beta}_0$ with $\boldsymbol{\beta}_0 \in \mathcal{N}(\mathbf{X})$ (i.\,e. $\mathbf{X} \boldsymbol{\beta}_0 = \mathbf{0}$) and $\boldsymbol{\beta}_1$ orthogonal to $\mathcal{N}(\mathbf{X})$.
Then
\begin{align}
  & \|\mathbf{y} - \mathbf{X}(\boldsymbol{\beta}_1 + \boldsymbol{\beta}_0)\|_2^2 + \gamma \| \boldsymbol{\beta}_1 + \boldsymbol{\beta}_0 \|_2^2 \notag \\
  & \qquad =
    \|\mathbf{y} - \mathbf{X} \boldsymbol{\beta}_1\|_2^2 + \gamma \| \boldsymbol{\beta}_1 \|_2^2 + \gamma \| \boldsymbol{\beta}_0 \|_2^2 \notag \\
  & \qquad \qquad \geq
    \|\mathbf{y} - \mathbf{X} \boldsymbol{\beta}_1\|_2^2 + \gamma \| \boldsymbol{\beta}_1 \|_2^2
\end{align}
with equality if and only if $\boldsymbol{\beta}_0 = \mathbf{0}$.
Thus the optimal $\hat{\boldsymbol{\beta}}$ is always orthogonal to the nullspace of $\mathbf{X}$.

\paragraph{Principal Components Regression}
The PCR coefficient vector can be calculated by
\begin{align}
  \boldsymbol{\beta}^{\mathrm{PCR}}_M = \sum_{m=1}^M \hat \theta_m \mathbf{v}_m
\end{align}
where $\mathbf{v}_m$ are the first $M$ right singular vectors of $\mathbf{X}$, thus the first $M$ columns of $\mathbf{V}_1$ \cite{hastie2009elements}.
The actual values of the coefficients $\hat \theta_m$ given by PCR are not necessary to show the orthogonality of $\boldsymbol{\beta}^{\mathrm{PCR}}_M$ to $\mathcal{N}(\mathbf{X})$:
\begin{align}
  \mathbf{V}_0^{\top} \boldsymbol{\beta}^{\mathrm{PCR}}_M = \sum_{m=1}^M \hat \theta_m \mathbf{V}_0^{\top} \mathbf{v}_m = \mathbf{0}\;.
\end{align}

\paragraph{Partial Least Squares}
This analysis is based on the recursive PLS algorithm from \cite{hastie2009elements}.
The algorithm 
starts with $\mathbf{X}_0 := \mathbf{X}$ and $\hat{\mathbf{y}}^{(0)} := \mathbf{0}$, where the data were assumed to be 
centered. In contrast to \cite{hastie2009elements}, we are using matrix notation instead of explicit sums. One recursion step $m \in\{ 1, \ldots, M \}$ consists of calculating a regression input 
\begin{align}
  \mathbf{z}_m = \mathbf{X}_{m-1} \mathbf{X}_{m-1}^{\top} \mathbf{y}
  \label{eq:PLS_zm}
\end{align}
and performing a univariate regression of $\mathbf{y}$ onto $\mathbf{z}_m$, giving $\hat \theta_m$ (the actual value is again not needed in this analysis), leading to an update for the estimated output
\begin{align*}
  \hat{\mathbf{y}}^{(m)} = \hat{\mathbf{y}}^{(m-1)} + \hat \theta_m \mathbf{z}_m\;.
\end{align*}
A recursion step ends by orthogonalizing the columns of $\mathbf{X}_{m-1}$ with respect to $\mathbf{z}_m$,
\begin{align}
  \mathbf{X}_{m} = \left(\mathbf{I} - \frac{1}{\mathbf{z}_m^{\top} \mathbf{z}_m}  \mathbf{z}_m \mathbf{z}_m^{\top} \right) \! \mathbf{X}_{m - 1}\;.
  \label{eq:PLS_Xm}
\end{align}
The PLS solution after $M$ steps is
\begin{align}
  \hat{\mathbf{y}}^{(M)} = \sum_{m=1}^M \hat \theta_m \mathbf{z}_m\;.
\end{align}
We show below that each $\mathbf{z}_m$ can be expressed in the form $\mathbf{z}_m = \mathbf{X} \mathbf{X}^{\top} \boldsymbol{\zeta}_m$ with some $\boldsymbol{\zeta}_m$.
Using this,
\begin{align}
  \hat{\mathbf{y}}^{(M)} = \mathbf{X}  \sum_{m=1}^M \underbrace{\hat \theta_m \mathbf{X}^{\top} \boldsymbol{\zeta}_m}_{\hat{\boldsymbol{\beta}}_m} =: \mathbf{X}  \boldsymbol{\beta}^{\mathrm{PLS}}_M
\end{align}
defines the regression coefficient $\boldsymbol{\beta}^{\mathrm{PLS}}_M$ and furthermore,
\begin{align}
  \mathbf{V}_0^{\top} \boldsymbol{\beta}^{\mathrm{PLS}}_M = \sum_{m=1}^M \mathbf{V}_0^{\top} \hat{\boldsymbol{\beta}}_m = \sum_{m=1}^M \hat \theta_m\mathbf{V}_0^{\top} \mathbf{X}^{\top} \boldsymbol{\zeta}_m = \sum_{m=1}^M \hat \theta_m\mathbf{V}_0^{\top} \mathbf{V}_1 \boldsymbol{\Sigma}_1 \mathbf{U}_1^{\top} \boldsymbol{\zeta}_m = \mathbf{0}\; ;
\end{align}
thus $\boldsymbol{\beta}^{\mathrm{PLS}}_M$ is orthogonal to $\mathcal{N}(\mathbf{X})$.
To show that $\mathbf{z}_m = \mathbf{X} \mathbf{X}^{\top} \boldsymbol{\zeta}_m$, insert \eqref{eq:PLS_zm} in \eqref{eq:PLS_Xm} to give
\begin{align}
  \mathbf{X}_{m}
    & = \left(\mathbf{I} - \frac{1}{\mathbf{z}_m^{\top} \mathbf{z}_m} \mathbf{X}_{m-1} \mathbf{X}_{m-1}^{\top} \mathbf{y} \mathbf{y}^{\top} \mathbf{X}_{m-1} \mathbf{X}_{m-1}^{\top} \right) \!\mathbf{X}_{m - 1} \notag \\
    & = \mathbf{X}_{m-1}\! \left( \mathbf{I} - \frac{1}{\mathbf{z}_m^{\top} \mathbf{z}_m} \mathbf{X}_{m-1}^{\top}  \mathbf{y} \mathbf{y}^{\top} \mathbf{X}_{m-1} \mathbf{X}_{m-1}^{\top} \mathbf{X}_{m-1} \right),
\end{align}
which implies that the product $\mathbf{X}_m \mathbf{X}_m^{\top}$ can be written with $\mathbf{X}_{m-1} \mathbf{X}_{m-1}^{\top}$ as a left factor,
\begin{align}
  \mathbf{X}_m \mathbf{X}_m^{\top} = \mathbf{X}_{m-1} \mathbf{X}_{m-1}^{\top} ( \cdots\!\,) \;.
\end{align}
This expression can be applied recursively,
\begin{align}
  \mathbf{X}_m \mathbf{X}_m^{\top} = \mathbf{X}_{m-1} \mathbf{X}_{m-1}^{\top} ( \cdots\!\,) = \mathbf{X}_{m-2} \mathbf{X}_{m-2}^{\top} ( \cdots\!\,) = \mathbf{X}_0 \mathbf{X}_0^{\top} ( \cdots\!\,) \;,
\end{align}
and, as $\mathbf{X}_0 = \mathbf{X}$,  $\mathbf{z}_{m + 1}$ can always be written as
\begin{align}
  \mathbf{z}_{m+1} = \mathbf{X} \mathbf{X}^{\top} \boldsymbol{\zeta}_{m+1}\;.
\end{align}

\section{Derivation of the Nullspace Projection}
\label{SI:derivation_projection}
The optimization 
\begin{subequations}
    \begin{alignat}{2}
        &\!\min_{\mathbf{v}}  \ \normx[2]{
\boldsymbol\beta_\Delta +\mathbf{v}}^2  \\
        &\text{subject to } \mathbf{Xv} =\mathbf{0},
    \end{alignat}
\end{subequations}
is a convex quadratic program with linear constraints. Its solution can be derived by introducing Lagrange multipliers to give the equivalent unconstrained optimization 
\begin{equation}
\min_{\mathbf{v},\mathbf{\lambda}} \ 
\tfrac{1}{2}
(\boldsymbol\beta_\Delta+\mathbf{v})^\top 
           (\boldsymbol\beta_\Delta +\mathbf{v}) + \mathbf{\lambda}^\top\mathbf{ Xv}.
\end{equation}
Set the derivatives of the objective function to zero to give
\begin{align}
\mathbf{v}^* &= -\boldsymbol\beta_\Delta - \mathbf{X}^\top \mathbf{\lambda}\nonumber \\
\mathbf{X}\mathbf{v}^* &= -\mathbf{X}\boldsymbol\beta_\Delta - \mathbf{X}\mathbf{X}^\top \mathbf{\lambda}=\mathbf{0} \nonumber\\
\mathbf{\lambda} &= -(\mathbf{X}\mathbf{X}^\top)^{-1}\mathbf{X}\boldsymbol\beta_\Delta \nonumber \\
\mathbf{v}^* &= (\mathbf{X}^\top(\mathbf{X}\mathbf{X}^\top)^{-1}\mathbf{X}-\mathbf{I})\boldsymbol\beta_\Delta
\end{align}

\section{Parabolic Data Example Predictions}
\label{si:parab_prediction_results}

\begin{figure}[H]
\centering
\begin{subfigure}{.47\linewidth}
  \centering
  \includegraphics[width=.95\linewidth]{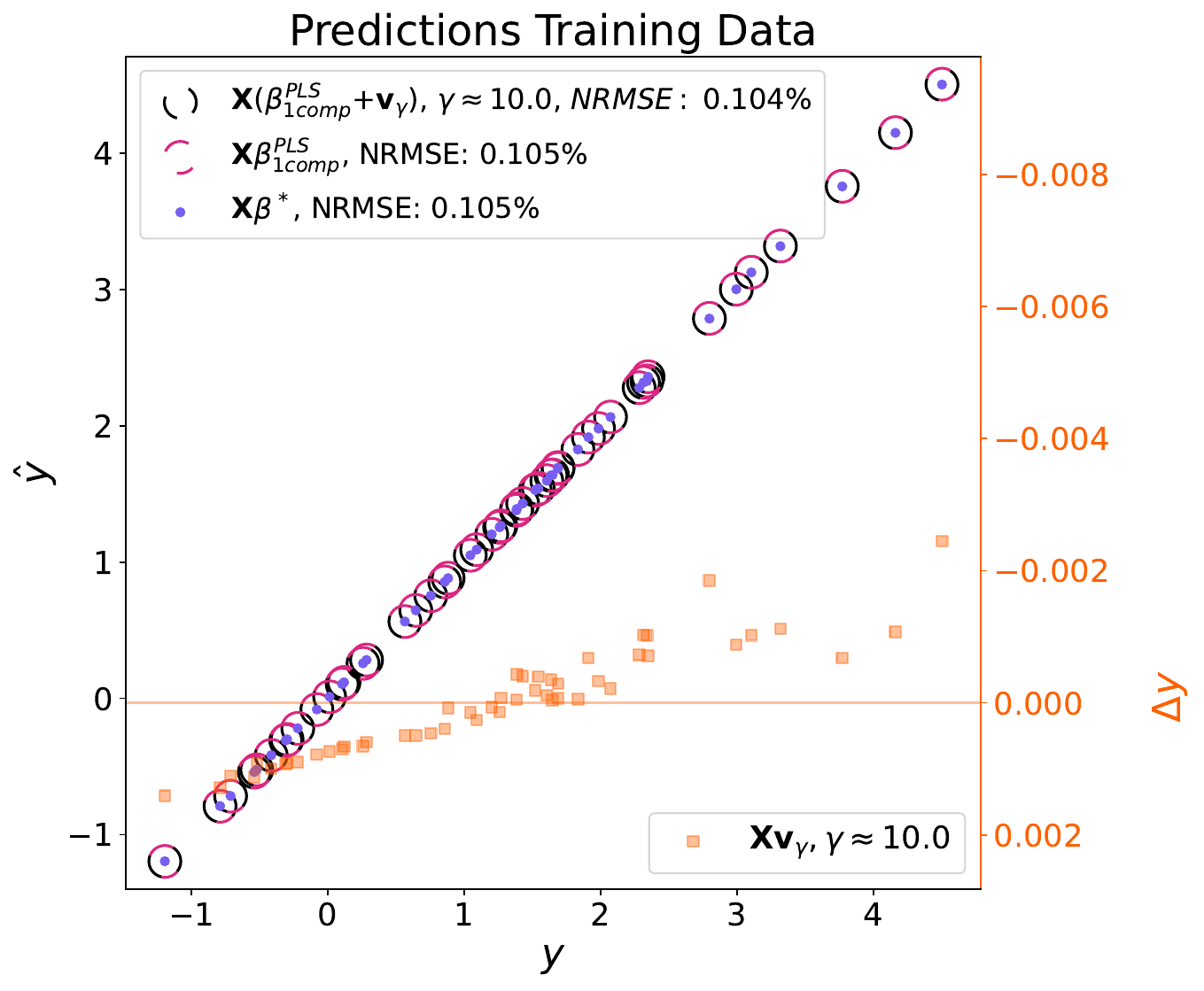}
  \caption{Predictions associated with the coefficients in Fig.\ \ref{fig:parab_ex}b.}
  \label{fig:scatter_pred_pls_betastar}
\end{subfigure}
\begin{subfigure}{.47\linewidth}
  \centering
  \includegraphics[width=.95\linewidth]{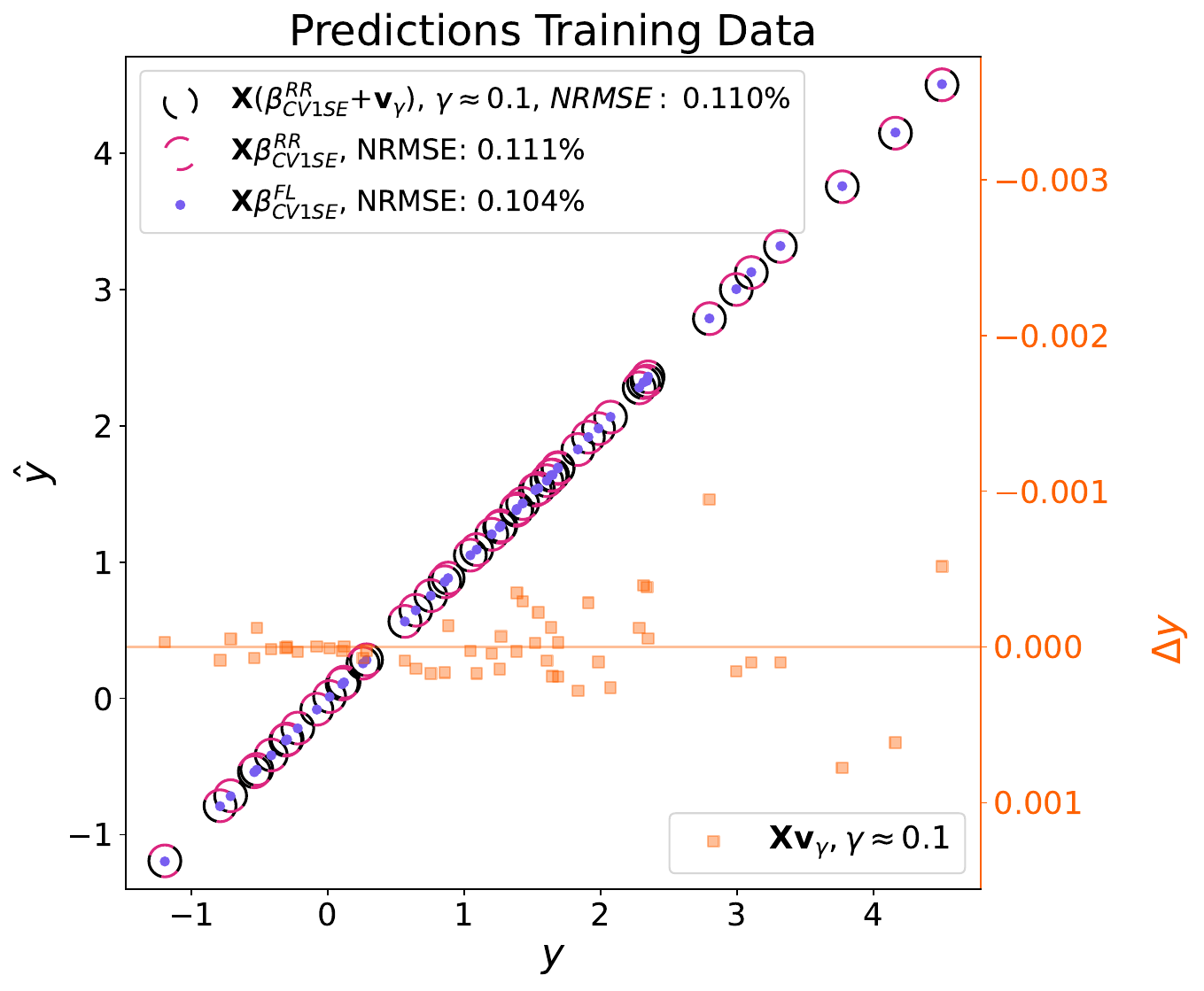}
  \caption{Predictions associated with the coefficients in Fig.\ \ref{fig:parab_ex_fl_rr}.}
  \label{fig:scatter_pred_rr_fl}
\end{subfigure}
\caption{Scatter plots of predictions associated with the parabolic data example.}
\label{fig:scatters_parab_predicitons}
\end{figure}

\section{Signal-to-Noise Ratio Approximation}
\label{si:snr_lfp}
\begin{figure}[H]
\begin{center}
    \includegraphics[width=.8\hsize]{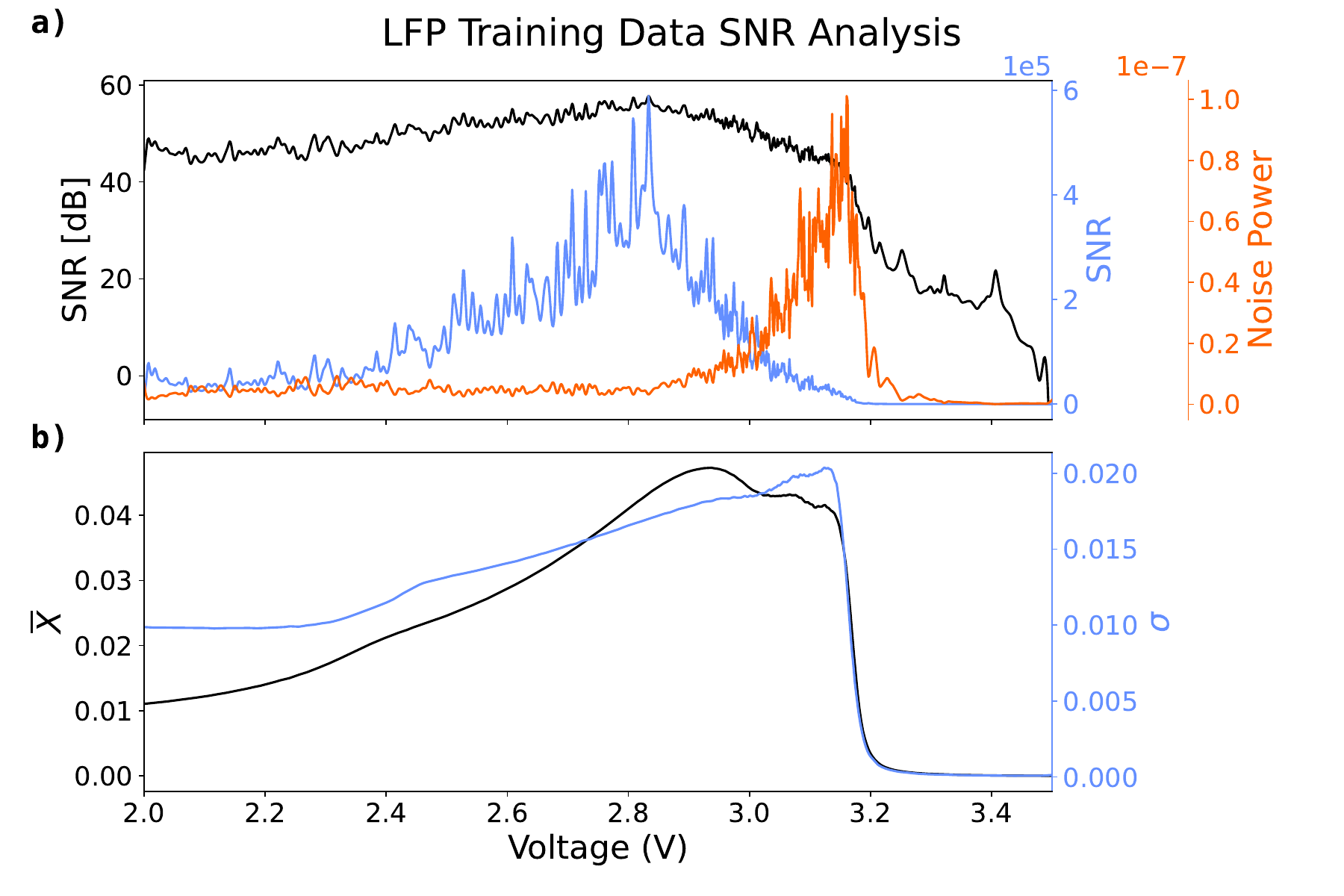}
\end{center}
\caption{\textbf{a)} Approximated \gls{snr} of the \gls{lfp} data set with \gls{snr} in dB, \gls{snr} ratio, and the noise power. \textbf{b)} Mean and standard deviation of the data.}
    \label{fig:SI_SNR_LLFP_Train}
\end{figure}
Figure \ref{fig:SI_SNR_LLFP_Train}a shows the approximated \gls{snr} of the \gls{lfp} data. The signal is estimated by fitting a spline, using \textit{scipy.interpolate.splrep} with a smoothing parameter $s=10^{-6}$ and the polynomial degree $k=3$, to the original data. The deviation to the spline is considered noise. %
The spline parameter choice depends on the expected degree of smoothness of the latent function. Figure \ref{fig:SI_SNR_LLFP_Train}b shows the associated mean and standard deviation. The \gls{snr} decreases strongly in the region 3.2--3.5\,V; however, in this region, the standard deviation of the data is also very low. Rescaling the data matrix columns to unit variance thus amplifies the noise in this section. 

The above analysis is only based on the data $\mathbf{X}$ without considering $\mathbf{y}$. Although standardization can amplify noise, whether a model based on z-scoring the data yields a higher prediction accuracy also depends on the relationship between $\mathbf{X}$ and $\mathbf{y}$. An alternative to mitigate the amplification of noise would be to rescale the data such that the variance of the column matches the normalized \gls{snr} ratio.

\printbibliography[heading=subbibliography]
\end{refsection}